\documentclass[lettersize,journal]{IEEEtran}
\usepackage{amsmath,amsfonts}
\usepackage{algorithmic}
\usepackage{algorithm}
\usepackage{array}
\usepackage{textcomp}
\usepackage{stfloats}
\usepackage{url}
\usepackage{verbatim}
\usepackage{graphicx}
\usepackage{cite}
\usepackage{multirow}
\usepackage{color,xcolor}
\usepackage{array}
\usepackage{booktabs}
\usepackage{epsfig}
\usepackage{amssymb}
\usepackage{graphicx}
\usepackage{subcaption}
\newif\ifshowcomments
\showcommentstrue

\ifshowcomments
\newcommand{\TMM}[1]{{\color{red}{[TMM: #1]}}}
\newcommand{\ACMMM}[1]{{\color[rgb]{0.2,0.7,0.2}{[ACMMM: #1]}}}
\else
\newcommand{\TMM}[1]{}
\newcommand{\ACMMM}[1]{}
\fi

\begin{document}

\title{Learning Physical-Spatio-Temporal Features for Video Shadow Removal}

\author{Zhihao Chen, Liang Wan$^\dag$ \textit{Member, IEEE}\thanks{Z. Chen, L. Wang$^\dag$, and Y. Xiao are with the College of Intelligence and Computing, Tianjin University, Tianjin 300350, China. Email: zh\_chen@tju.edu.cn, lwan@tju.edu.cn, fnxyf@tju.edu.cn.}, Yefan Xiao, Lei Zhu \textit{Member, IEEE}\thanks{L. Zhu is with The Hong Kong University of Science and Technology (Guangzhou), Nansha, Guangzhou, 511400, Guangdong, China and The Hong Kong University of Science and Technology, Hong Kong SAR, China  (Email: leizhu@ust.hk) }, Huazhu Fu \textit{Senior Member, IEEE}\thanks{H. Fu is with the Institute of High Performance Computing (IHPC), Agency for Science, Technology and Research (A*STAR), Singapore 138632. (E-mail: hzfu@ieee.org)}\\
}

\markboth{IN SUBMISSION}%
{Shell \MakeLowercase{\textit{et al.}}: A Sample Article Using IEEEtran.cls for IEEE Journals}


\maketitle

\begin{abstract}
Shadow removal in a single image has received increasing attention in recent years. However, removing shadows over dynamic scenes remains largely under-explored.  In this paper, we propose the first data-driven video shadow removal model, termed PSTNet, by exploiting three essential characteristics of video shadows, i.e., physical property, spatio relation, and temporal coherence. Specifically, a dedicated physical branch was established to conduct local illumination estimation, which is more applicable for scenes with complex lighting and textures, and then enhance the physical features via a mask-guided attention strategy. Then, we develop a progressive aggregation module to enhance the spatio and temporal characteristics of features maps, and effectively integrate the three kinds of features. Furthermore, to tackle the lack of datasets of paired shadow videos, we synthesize a dataset (SVSRD-85) with aid of the popular game GTAV by controlling the switch of the shadow renderer. Experiments against 9 state-of-the-art models, including image shadow removers and image/video restoration methods, show that our method improves the best SOTA in terms of RMSE error for the shadow area by $14.7\%$. In addition, we develop a lightweight model adaptation strategy to make our synthetic-driven model effective in real world scenes. The visual comparison on the public SBU-TimeLapse dataset verifies the generalization ability of our model in real scenes.
\end{abstract}


\section{Introduction}
\label{sec::introduction}
Shadows, created wherever an object obscures the light source, are one ubiquitous natural phenomenon in our daily life. 
They can provide useful hints for extracting scene geometry~\cite{okabe2009attached,karsch2011rendering}, light direction~\cite{lalonde2009estimating}, and camera location and its parameters~\cite{junejo2008estimating}. 
However, shadows can also hinder the performance of those downstream high-level computer vision tasks, such as object/change detection~\cite{surkutlawar2013shadow,muller2019brightness,su2016shadow}, face recognition~\cite{zhang2018improving}, due to spurious edges on the boundaries of shadows and confusion between albedo and shading.
The last two decades have witnessed a rapid development in image shadow removal.
Early works usually examine illumination and gradients priors~\cite{gryka2015learning,finlayson2005removal,shor2008shadow,yang2012shadow}. 
With the aids of the datasets of the real world shadow and shadow-free images, researchers developed many data-driven CNN-based approaches, which are learnt in an end-to-end manner~\cite{qu2017deshadownet,wang2018stacked,hu2019direction,cun2020towards}. In recent years, several research works combine the physical priors and CNN models to transform the shadow removal task to an image decomposition issue~\cite{le2021physics,fu2021auto,le2020shadow}.

In striking contrast with the flourishing development of image shadow removal, few works have been explored in shadow removal over dynamic scenes. 
As one type of video restoration tasks, 
video shadow removal can benefit various downstream high-level tasks, such as dynamic scene understanding~\cite{jung2009efficient,wang2009real} and object tracking~\cite{nadimi2004physical}. 
On the other hand, we also notice that many other video restoration tasks have received significant research interests in recent years, including video super-resolution\cite{chan2021basicvsr,isobe2020video1,isobe2020video2}, video deblur~\cite{kim2017dynamic,zhang2018adversarial}, video denoise~\cite{maggioni2021efficient,yue2020supervised}, video derain~\cite{yue2021semi,kim2015video}, etc.
Hence, it is of great importance to address the problem of video shadow removal. 
The brute-force solution is directly using existing image shadow removers in videos, however, it may ignore the potential effect that temporal information can bring, and is prone to generate unstable predictions between adjacent video frames~\cite{le2021physics}.

In this work, we propose a novel video shadow removal method, termed \textbf{PSTNet}, by exploiting three main characteristics related to shadows, i.e. physical, spatio, and temporal characteristics, simultaneously.
\textcircled{1} Physical characteristic depicts the shadows by a simplified linear illumination model. Some ralated works~\cite{le2021physics, le2020shadow} pointed out that employing the physical characteristic of shadows can make shadow removal mapping easier to constrain.
\textcircled{2} Spatio characteristic depicts the position-sensitive semantic information of shadows, which reflects fine textures in the image.
\textcircled{3} Temporal characteristic depicts the motion condition of shadows. Other video restoration tasks~\cite{wang2019edvr,chan2021basicvsr} have shown that the temporal-continuity knowledge is able to facilitate the prediction for the current frame given continuous frames.
To be specific, we use three parallel branches to extract the embedded features from physical, spatio, and temporal characteristics, respectively. 
Physical branch model the shadow removal problem as the over-exposure problem which represented by the linear regression function, following SID~\cite{le2021physics}. To overcome the inability of SID to solve complex lighting and texturing problems, we equip the physical branch with an adaptive exposure estimation module, which considers exposure bias in different shadow regions. 
Meanwhile, a mask-guided supervised attention module is leveraged to guide our model pay more attention to the features in shadow regions rather than non-shadow regions. 
In addition, to maintain the spatial resolution and temporal coherence of the prediction results, we introduce the spatio and temporal characteristic branches, respectively. 
Finally, we design a dedicated feature fusion module to aggregate those features to refine the final shadow removal results in a progressive manner.

It is worth pointing out there is no dataset with paired shadow and shadow-free videos, since it is almost impossible to reproduce dynamics exactly between two shots in real-world scenes. 
Hence, in this work, we synthesize one video shadow removal dataset, termed \textbf{SVSRD-85}, for training and evaluating our model.
SVSRD-85 collects paired shadow and shadow-free videos in the popular game GTAV~\cite{GTAV2015} by controlling the switch of shadow renderer.
It contains 85 videos with 4250 frames, covering different object categories and  vairous motion/illumination conditions, while 
all the frames have their corresponding high-quality shadow-free ground-truth. 
We hope that the release of this dataset will facilitate research on video shadow removal. To demonstrate the effectiveness of our method, we present a comprehensive evaluation against 9 state-of-the-art related models on our SVSRD-85 dataset. Results show that our model significantly outperforms existing methods, including single image shadow removers~\cite{guo2012paired,gong2014interactive, hu2019direction,le2021physics, cun2020towards,fu2021auto}, image restoration methods~\cite{zamir2021multi}, and video restoration methods~\cite{wang2019edvr,chan2021basicvsr}.

Finally, to make the model trained in synthetic data effective in a real-world scenes, we propose a lightweight model transformation strategy, termed S2R, to adapt the synthetic-driven models to the real world scenes without retraining. Visualization Comparison proves the effectiveness of our method.


\section{Related Works}
\label{sec:relatedworks}
\subsection{Image Shadow Removal}
Early research on image shadow removal employ prior information, e.g., gradient~\cite{gryka2015learning}, illumination~\cite{shor2008shadow,xiao2013fast,zhang2015shadow}, and region~\cite{guo2012paired,vicente2017leave}, for removing shadows. 
However, conventional prior
based methods can just handle high-quality images with certain constraints of real scenes (e.g. ample lighting conditions and simple textures).
In recent years, deep learning based methods boost the removal performance~\cite{qu2017deshadownet,ding2019argan,hu2019direction,cun2020towards}, since CNN networks have built-in inductive biases that make them well-suited to a wide variety of computer vision applications~\cite{liu2022convnet}. 
In more detail, DeshadowNet~\cite{qu2017deshadownet} proposes a multi-context architecture, where the  shadow matte is predicted by embedding information from three different perspectives, i.e. global localization, appearance modeling and semantic modeling.
ST-CGAN~\cite{wang2018stacked} leverages two stacked conditional GANs to jointly train shadow detection and removal tasks.
DSC\cite{hu2019direction} designs direction-aware context to improve shadow detection and removal.
DHAN~\cite{cun2020towards} analyses two types of shadow ghosts and proposes a dual hierarchical aggregation network and shadow augmentation method to boost shadow removal.
Zhu et al.~\cite{zhu2022bijective} argued that shadow removal and generation are interrelated. They proposed BMNet, which couples the learning procedures of shadow removal and shadow generation in a unified parameter-shared framework.

In the literature, some methods consider physical models of shadow formation and embed such information in the end-to-end deep learning manner. 
SID~\cite{le2021physics} uses a linear illumination transformation to model the shadow effects in the image, which expresses the shadow image as a combination of the shadow-free image, the shadow parameters, and a matte layer. It then uses two deep networks to predict the shadow parameters and the shadow matte respectively.
Fu et al.~\cite{fu2021auto} extended SID via employing multi-exposure image fusion strategy and proposes a boundary-aware RefineNet to eliminate the remaining ghost shadow further. 

In addition, to alleviate the requirements of capturing paired data, some GAN based methods~\cite{hu2019mask,liu2021LG,liu2021G2R} perform image shadow removal on unpaired shadow and shadow-free images.
MaskShadowGAN~\cite{hu2019mask} learns to produce a shadow mask from the input shadow image and then takes the mask to guide the shadow generation via re-formulated cycle-consistency constraints. LG-ShadowNet~\cite{liu2021LG} improves MaskShadowGAN by introducing a brightness guidance strategy. G2R-ShadowNet~\cite{liu2021G2R} generates unpaired data given a set of shadow images and their corresponding shadow masks. Gao et al.~\cite{gao2022towards} proposed a shadow simulation method to simulate shadow on the grayscale, which can be applied to arbitrary shadow-free images and masks to generate corresponding shadow images.
However, due to the lack of accurate pixel-level shadow-free supervision, those unpaired-based methods still suffer from artifacts and image blur, and hence there exists a certain gap in performance compared to the fully supervised approaches on benchmarks.

\subsection{Video Shadow Removal}
Video shadow removal aims to remove shadows from each frame of a video. Existing methods almost all rely on hand-crafted features and there is no deep-learning based method for video shadow removal. 
Existing methods usually assume that the background is relatively stationary, allowing the moving objects and shadows to be separated. 
Nadimi \& Bhanu~\cite{nadimi2004physical} separated moving cast shadows from the moving objects in an outdoor environment. This approach is based on a spatio-temporal albedo test and dichromatic reflection model which accounts for both the sun and the sky illuminations. 
Jung et al.~\cite{jung2009efficient} presented a statistical method for background subtraction and shadow removal for grayscale video sequences. 
Wang et al.~\cite{wang2009real} presented an approach of moving vehicle detection and cast shadow removal for video based traffic monitoring. Based on conditional random field, spatial and temporal dependencies in traffic scenes are formulated under a probabilistic discriminative framework.

The above methods are suitable for surveillance videos with static backgrounds, but not applicable to remove shadows in scenes with rich motion conditions. In this paper, we develop the first deep-learning based method for video shadow removal. Furthermore, we  bypass the strict requirements of collecting paired videos from real scenes, and built one virtual dataset of paired shadow and shadow-free videos in synthetic scenes.



\section{SVSRD-85: Synthetic Video Shadow Removal Dataset}
\label{sec:dataset}
\begin{figure*}
\centering
\includegraphics[width=\textwidth]{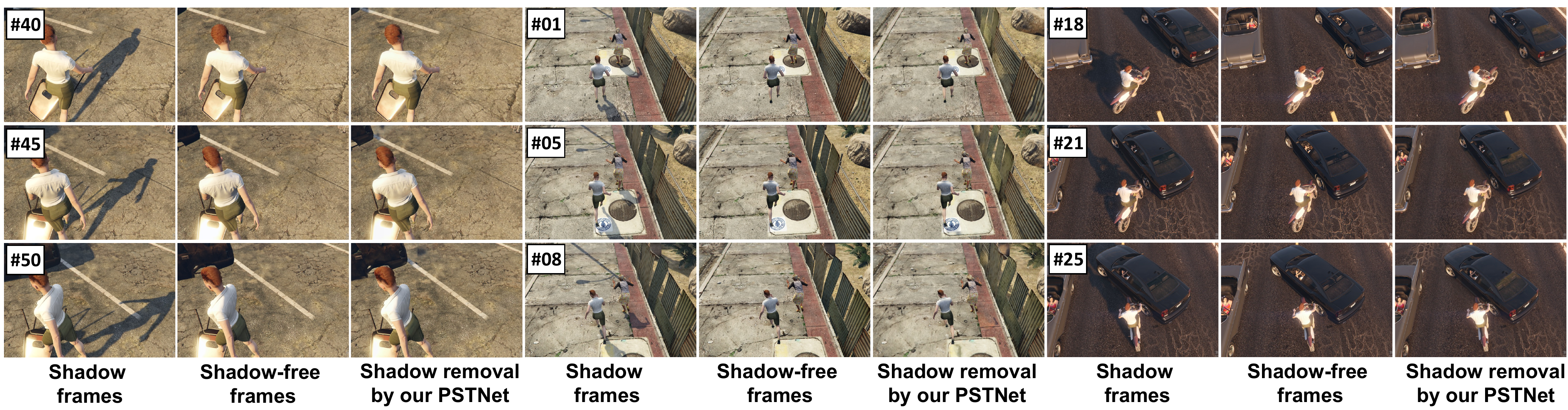}
\caption{Example frames of our  synthetic dataset (SVSRD-85), and video shadow removal results by our proposed PSTNet. }
\label{fig:dataset}
\vspace{-2.5mm}
\end{figure*}
\noindent
\textbf{Motivation.}
In real world, though we can easily get a shadowed video, getting the synchronous shadow-free video is quite difficult since it's hard to restore the same shooting states between two shots. 
Le~\textit{et al.} collected SBU-Timelapse~\cite{le2021physics}, a video dataset of 50 videos with time-lapse photography, in which each video contains a static scene without visible moving objects.
They used ``max-min'' technique to obtain a single pseudo shadow-free frame for each video.
However, this dataset has two deficiencies. First, the obtained pseudo shadow-free image still contains much static shadows (as shown in Figure~\ref{fig:dataset}). 
Second, it does not allow for any movement of targets and perspectives, which greatly limits the motion richness.
\renewcommand\arraystretch{1.1}
\begin{table}[t]
\begin{center}
  \caption{SVSRD-85 vs. SBU-TimeLapse~\cite{le2021physics}.}
  \vskip -5pt
  \label{table:SVSRD_vs_SBUTimeLapse}
  \resizebox{0.46\textwidth}{!}{%
    \begin{tabular}{c|c|c|c|c}
        \toprule[1pt]
        Dataset & Type & Annotation & Scenario & \#Videos\\
        \specialrule{0em}{1pt}{1pt}
        \hline
        \specialrule{0em}{1pt}{1pt}
        SVSRD-85 & Synthetic & Paired & Dynamic & 85  \\
        SBU-TimeLapse~\cite{le2021physics} & Real & Unpaired & Static & 50  \\
        \bottomrule[1pt]
    \end{tabular}
    }
    \vspace{-5mm}
  \end{center}
\end{table}

\noindent
\textbf{Synthesize dataset from game scenes.}
In order to get the reliable video pairs to train, we consider rendering shadow and shadow-free video pairs in synthetic scenes.
Thanks to the strong physical engine and the editing flexibility of the popular game: Grand Theft Auto V (GTAV), many works~\cite{richter2016playing,sidorov2019conditional} explored to drive various computer vision tasks based on this game. GTAV provides a range of APIs that can monitor the creation, modification, and deletion of resources used to specify the scene and synthesize images/videos. 
What's more, it supports to control the switch of the shadow renderer, which can allow us to obtain the high-quality video shadow pairs.

In this paper, we collect a synthetic video shadow removal dataset (SVSRD-85) via the above manner. 
Our dataset includes 85 videos with total 4250 frames, and each video contains a sequence of shadow images and corresponding shadow-free ground-truth. 
We compute the pesudo shadow masks by operating Otsu’s algorithm to the difference between shadow and shadow-free images, similar to MaskShadow-GAN~\cite{hu2019mask}. 
To provide guidelines for future works, we randomly split the dataset into training and testing sets with a ratio of 7:3, then obtain 59 training videos and 26 testing videos. 
It is worth noting that since there are no restrictions on shooting conditions, SVSRD-85 includes rich scenes with more complex illustration and surface textures, compared with SBU-Timelapse. Table~\ref{table:SVSRD_vs_SBUTimeLapse} shows the main differences between SVSRD-85 and SBU-Timelapse. Some examples of video in SVSRD-85 can be found in Figure~\ref{fig:dataset}. In this paper, we also use some videos in SBU-Timelapse to verify the generalisation ability of our model in real scenes.
\begin{figure*}[!t]
\centering
\includegraphics[width=1\textwidth]{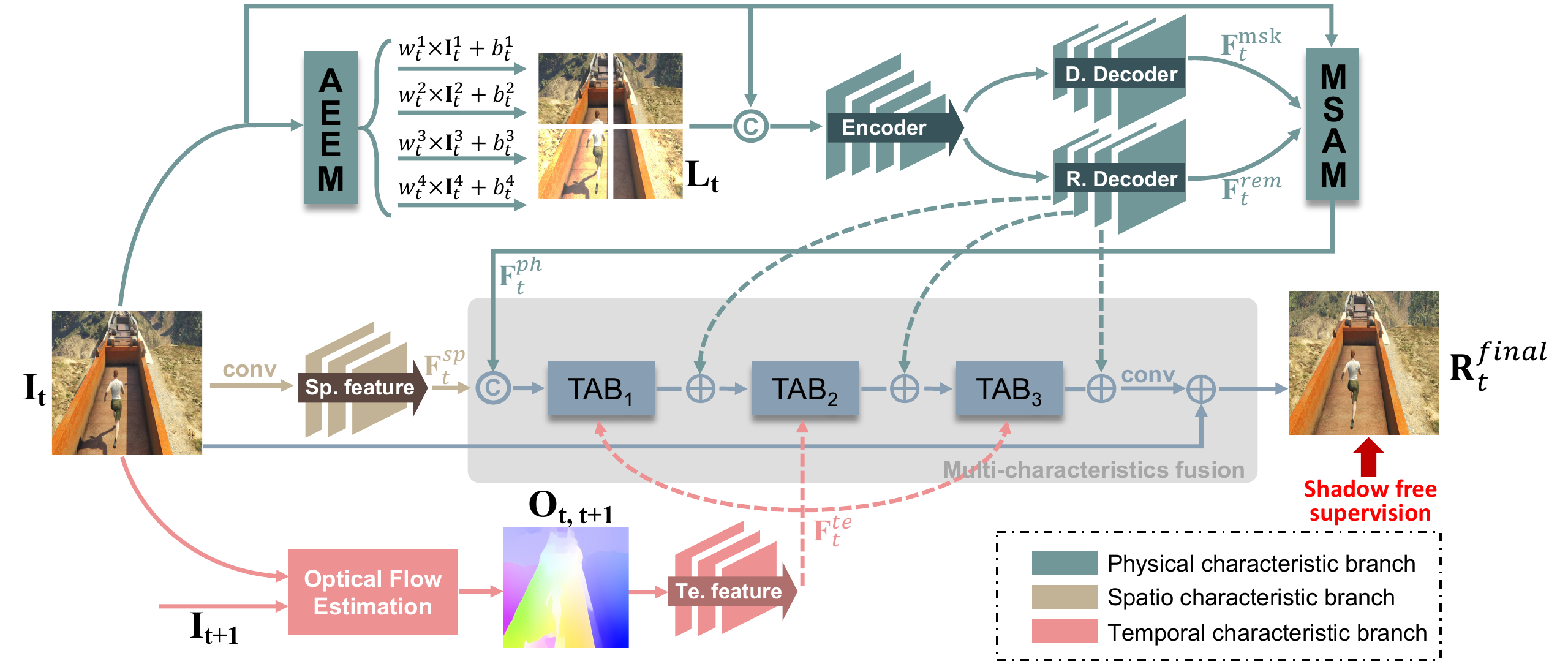}
\caption{The schematic illustration of our proposed PSTNet. The current frame $\mathbf{I}_{t}$ is fed into three parallel branches to extract features of three different characteristic: physical characteristic (\textcolor[RGB]{113,151,152}{top branch}), spatio characteristic (\textcolor[RGB]{159,179,150}{middle branch}), and temporal characteristic (\textcolor[RGB]{237,145,147}{bottom branch}). In bottom branch, the adjacent frame $\mathbf{I}_{t+1}$ will also be used to compute the optical flow $\mathbf{O}_{t, t+1}$. Next, a tailored feature fusion module will aggregate those multi-characteristic features in a progressive manner, thereby performing the final shadow removal prediction $\mathbf{R}_{t}^{final}$. }
\label{fig:framework}
\vspace{-3mm}
\end{figure*}
\section{Proposed Method} \label{sec:method}
To the best of our knowledge, so far there is no deep-learning based method for video shadow removal. In this paper, we propose a baseline network for this task, termed \textbf{PSTNet}, which learns the union features of physical, spatio, and temporal characteristics to perform video shadow removal. 

\subsection{Overview of Our Network}
\label{sec:overview}
Figure~\ref{fig:framework} presents the schematic illustration of our PSTNet. The network takes the frame $\mathbf{I}_{t}$ and its next frame $\mathbf{I}_{t+1}$ as inputs, then outputs the shadow removed result of the $t$-th frame in an end-to-end manner. 
The intuition behind our network is to leverage complementary information of physical, spatio, and temporal characteristics of moving shadows. 
We explicitly extract the three kinds of features with three independent branches, followed by a multi-characteristics fusion module to achieve the hybrid features in a progressive manner. For convenience, we first introduce the workflows of the three branches.

\noindent
\textbf{Physical characteristic branch.} Shadows are regular physical phenomena produced by occluded lighting. The removal of shadows is inseparable from the analysis of physical lighting conditions. 
Following SID~\cite{le2021physics}, shadow removal can be considered as the physical re-exposure problem. However, SID uses a uniform light model for different areas, ignoring the effects of complex lighting and background textures.
To remedy this problem, we extract the physical characteristics by estimating the adaptive exposure parameters in each shadow region, and using an encoder-decoder sub-network to further learn the broad contextual information due to large receptive fields. 

Specifically, the shadow image $\mathbf{I}_{t}$$\in$$\mathcal{R}^{3 \times W\times H}$ is first fed into AEEM (Adaptive Exposure Estimation Module) to estimate the re-exposure parameters, yielding the over-exposure image $\mathbf{L}_{t}$$\in$$\mathcal{R}^{3 \times W\times H}$. 
Then we feed the concatenation of $\mathbf{I}_{t}$ and $\mathbf{L}_{t}$ into a encoder-decoder sub-network to learn the physical features in the hierarchical manner. 
To make the physical feature more focused on the shadow regions, we append an extra  decoder served for shadow detection. Here, we utilize a widely-used UNet~\cite{UNet2015} structure as the encoder-decoder backbone. 
By this way, we can obtain $\mathbf{F}_{t}^{rem}$ and $\mathbf{F}_{t}^{msk}$ from shadow removal decoder and shadow mask decoder, respectively. 
To further enhance the physical features, we design a Mask-guided Supervised Attention Module (MSAM). 
It enhances the physical features $\mathbf{F}_{t}^{rem}$ to $\mathbf{F}_{t}^{ph}$ with the guidance of the ground-truth supervision through attention mechanism. 

\noindent
\textbf{Spatio characteristic branch.} Shadow removal is a position-sensitive task, since it needs to establish the pixel-to-pixel correspondence from the input to the output.
To preserve the desired fine texture in the final results, in the spatio branch, we employ a sub-network that operates on the original input frame resolution (without any downsampling operation).
To be specific, the sub-network extracts spatio features with a $3$$\times$$3$ convolution followed by a self channel attention block which has the same structure as Figure~\ref{fig:TAB} (b). Then we can get spatio features $\mathbf{F}_{t}^{sp}$. 

\noindent
\textbf{Temporal characteristic branch} is committed to extract the temporal knowledge for the current frame. 
In video restoration tasks~\cite{wang2019edvr, chan2021basicvsr}, the use of adjacent frames has been shown to facilitate the prediction of the current frame due to the continuity of the video. 
The temporal information brought by adjacent frames are apt to distinguishes different objects in the current frame.
Here, we model the temporal information by computing the optical flow map.
Specifically, the current frame $\mathbf{I}_{t}$ and the next frame $\mathbf{I}_{t+1}$ are fed into a well-trained optical flow estimation network (here, we choose FlowNet2~\cite{ilg2017flownet}) to generate the optical flow $\mathbf{O}_{t, t+1}$$\in$$\mathcal{R}^{3 \times W\times H}$. 
Next, $\mathbf{O}_{t, t+1}$ is further encoded by a $3$$\times$$3$ convolution followed by a self channel attention block which has the same structure as Figure~\ref{fig:TAB} (b), which yields the temporal features $\mathbf{F}_{t}^{te}$.

\begin{figure}[t]
\centering
\includegraphics[width=0.45\textwidth]{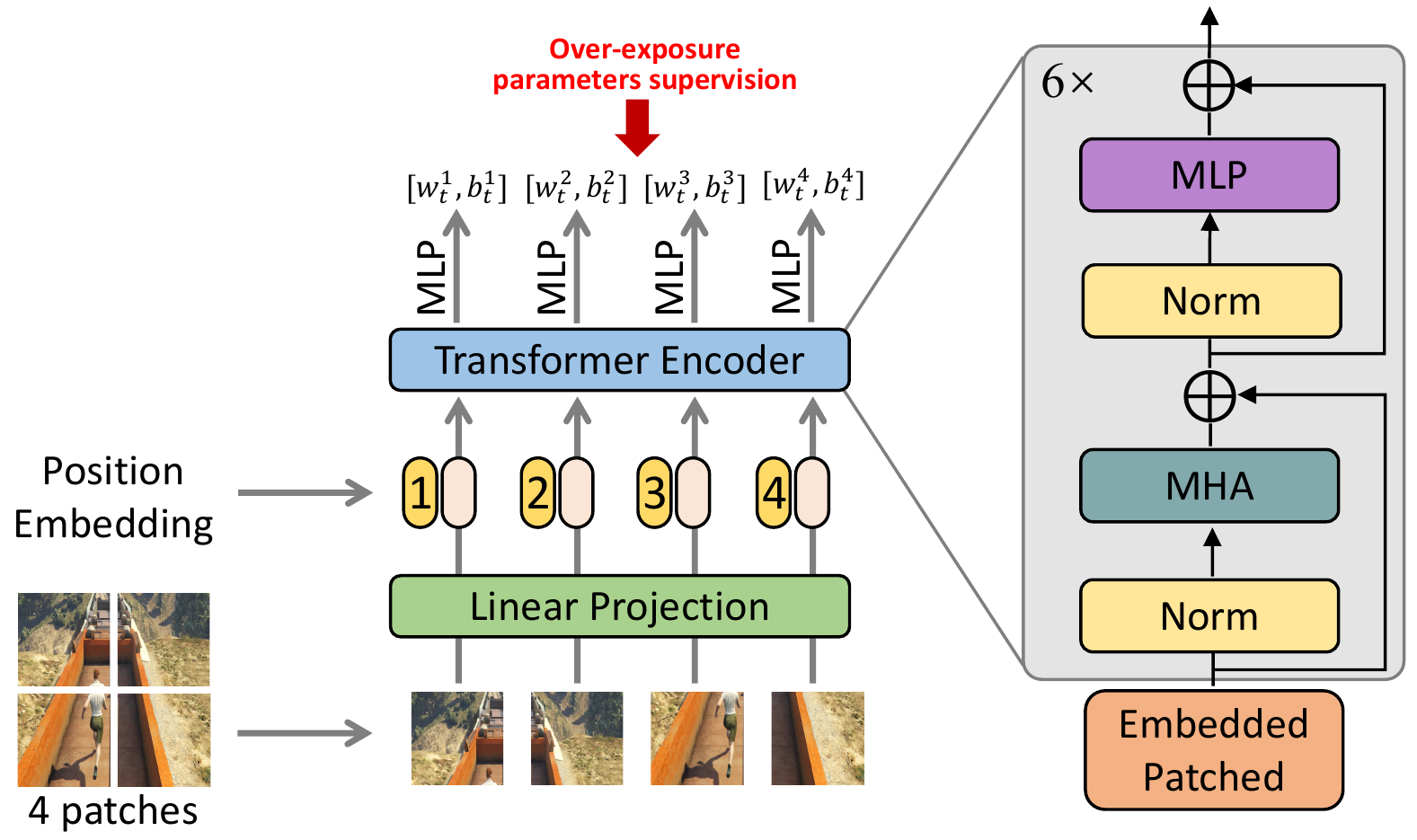}
\caption{The schematic illustration of our Adaptive Exposure Estimation Module (AEEM). }
\label{fig:AEEM}
\vspace{-2.5mm}
\end{figure}

\subsection{Adaptive Exposure Estimation Module}
\label{sec:aeem}
Considering a shadow frame $\mathbf{I}_{t}$, some recent shadow removal methods~\cite{le2021physics,fu2021auto}, which use physical shadow models, mainly learn to over-exposure $\mathbf{I}_{t}$ to a lightened version, and then fuse them to acquire the  shadow-free image by estimating a shadow matte. 
The common ground of these works is that for simplicity, only one group of linear parameters (denoted as $[\mathrm{w}, \mathrm{b}]$) is predicted for the entire frame $\mathbf{I}_{t}$. With estimated $[\mathrm{w}, \mathrm{b}]$, the over-exposure image $\mathbf{L}_{t}$ can be formulized as follow:
\begin{equation}\label{Equ:I2L}  
    \mathbf{L}_t = \mathrm{w}_t \times \mathbf{I}_t + \mathrm{b}_t.
\end{equation}
However, due to the complex changes in lighting and texture, it may be difficult to fit different regions with only one set of exposure parameters. 
In our work, we introduce the Adaptive Exposure Estimation Module (AEEM) to estimate respective $[\mathrm{w},\mathrm{b}]$ for different regions rather than the whole image. 
This can alleviate conflicts in case that  $[\mathrm{w},\mathrm{b}]$ for different regions are inconsistent. 
What's more, we utilize the transformer encoder to build the long-range relationships between image patches, which smooths the estimations of each $[\mathrm{w},\mathrm{b}]$. After AEEM, we can obtain the over-exposure image $\mathbf{L}_{t}$.

As illustrated in Figure~\ref{fig:AEEM}, AEEM takes the shadow image $\mathbf{I}_t$ as input and first splits it into $N$ patches. In our experiments, we set $N$=$4$. 
With patch embedding and position embedding, we employ a transformer encoder to produce the $[\mathrm{w},\mathrm{b}]$ sequence, denoted as $\{[\mathrm{w}_t^n, \mathrm{b}_t^n]\}_{n=1}^N$. 
For convenience, we inherit the same transformer encoder structure as the widely known ViT~\cite{dosovitskiy2020image}, which uses 6 transformer blocks with cascaded multi-head attention module and multi-layer perceptron.
After obtaining $\{[\mathrm{w}_t^n, \mathrm{b}_t^n]\}_{n=1}^N$, we re-exposure the shadow image $\mathbf{I}_t$ to a lightened version $\mathbf{L}_t$ by:
\begin{equation}\label{Equ:It2Lt}  
    \mathbf{L}_t^n = \mathrm{w}_t^n \times \mathbf{I}_t^n + \mathrm{b}_t^n,
\end{equation}
where $\mathbf{I}_t^n$ is $n$-th patch of $\mathbf{I}_t$; $\mathbf{L}_t^n$ is the corresponding lightened version of $\mathbf{}_t^n$. 
Then, we obtain the estimated $\mathbf{L}_t$ by assembling $\{\mathbf{L}_t^n\}^N_{n=1}$ in the original spatial order.
Compared with estimating uniform [w,b] for the whole image as SID\cite{le2021physics}, our proposed AEEM can be proved to handle more complex scenarios (see \ref{sec:ablation} for details).

\subsection{Mask-guided Supervised Attention Module}
\label{sec:msam}

Mask-guided Supervised Attention Module (MSAM) is employed to enhance the physical features $\mathbf{F}_{t}^{rem}$ to $\mathbf{F}_{t}^{ph}$ with the guidance of the ground-truth supervision through attention mechanism. 
The motivations behind MSAM include third main aspects. 
First, ground-truth supervisory signals is useful for progressive shadow removal. 
Second, a well-segmented shadow mask can guide the network to suppress the less informative features (mainly existing in non-shadow regions) and enhance the useful features (mainly existing in shadow regions).
Third, multi-task learning can leverage a stronger encoder via training with multiple types of supervised labels.

\begin{figure}[]
\centering
\includegraphics[width=0.48\textwidth]{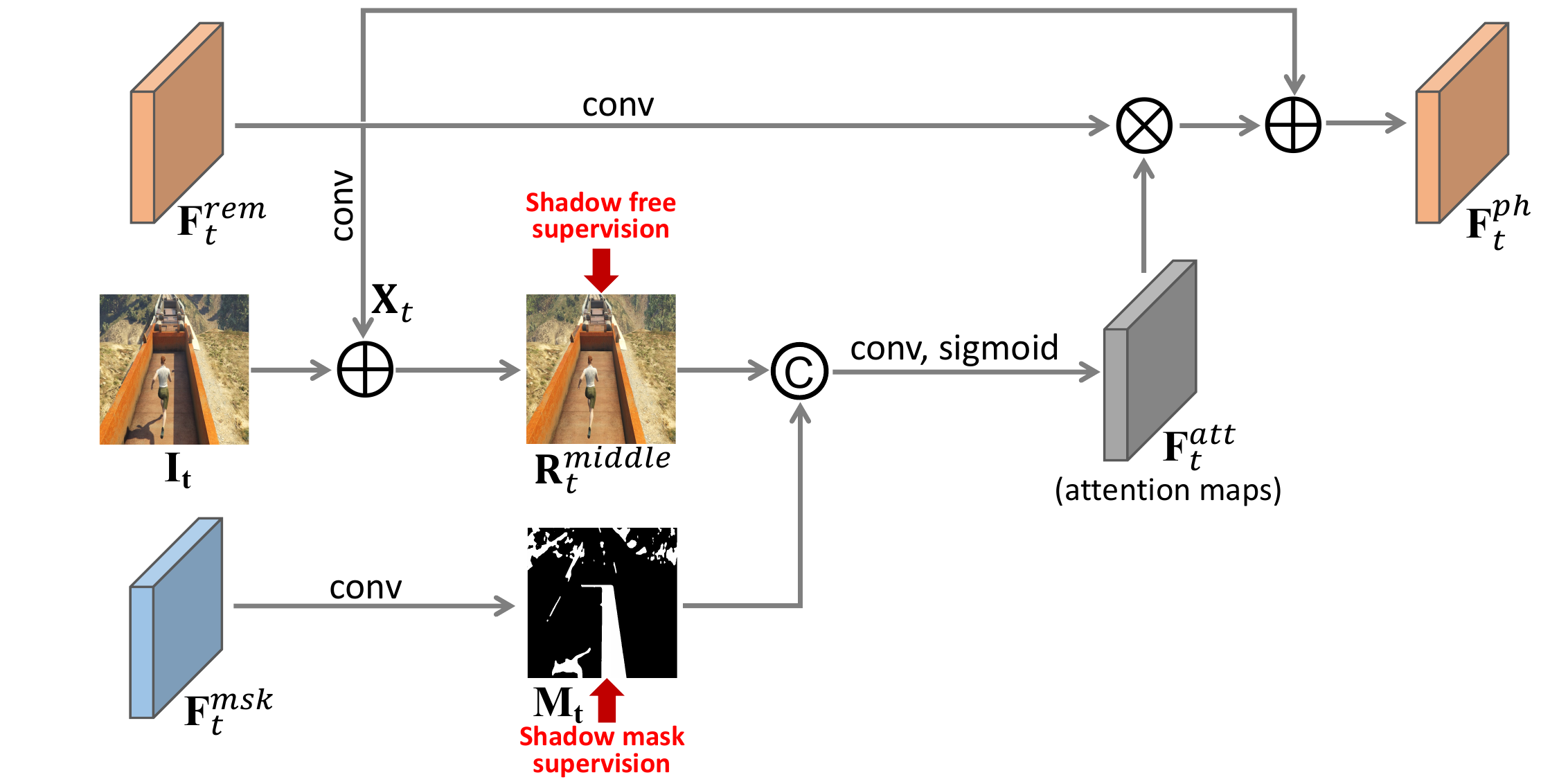}
\caption{The schematic illustration of  Mask-guided Supervised Attention Module (MSAM).}
\label{fig:MSAM}
\vspace{-2.5mm}
\end{figure}

As illustrated in Figure~\ref{fig:MSAM}, on the one hand, MSAM takes the features $\mathbf{F}_t^{rem}$$\in$$\mathcal{R}^{C \times W\times H}$ from shadow removal decoder to generate the residual image $\mathbf{X}_t$$\in$$\mathcal{R}^{3 \times W\times H}$ with a simple $1$$\times$$1$ convolution, where $W\times H$ denotes the spatial dimension and $C$ is the number of channels. 
The residual image $\mathbf{X}_t$ is added to the input image $\mathbf{I}_t$ to obtain the coarse shadow-free estimation $\mathbf{R}_t^{middle}$. 
\begin{equation}\label{Equ:R_middle}  
    \mathbf{R}_t^{middle} = \mathbf{I}_t + \mathtt{Conv}(\mathbf{F}_t^{rem}),
\end{equation}
where $\mathtt{Conv}$ denotes the simple $1$$\times$$1$ convolution. On the other hand, MSAM takes the feature $\mathbf{F}_t^{msk}$$\in$$\mathcal{R}^{C \times W\times H}$ from shadow detection decoder to generate the shadow mask prediction $\mathbf{M}_t$$\in$$\mathcal{R}^{1 \times W\times H}$. 
For these predicted image $\mathbf{R}_t^{middle}$ and $\mathbf{M}_t$, we exert explicit supervision with the ground-truth image. 
Next, per-pixel attention maps $\mathbf{F}_t^{att}$$\in$$\mathcal{R}^{C \times W\times H}$ are generated from the hybrid feature, concatenated by $\mathbf{R}_t^{middle}$ and $\mathbf{M}_t$, using a simple $1$$\times$$1$ convolution followed by the sigmoid activation.
Then, $\mathbf{F}_t^{att}$ are then employed to guide the transformed $\mathbf{F}_t^{rem}$ (obtained after $1$$\times$$1$ convolution), resulting in the attention-guided residual features which are added to the identity mapping path. Consequently, the attention-augmented $\mathbf{F}_t^{ph}$ can be formulized as follows:
\begin{equation}\label{Equ:F_ph}  
\begin{aligned}
    \mathbf{F}_t^{ph} &=& \mathbf{F}_t^{rem} + \mathtt{Conv}(\mathbf{F}_t^{rem}) \odot \mathbf{F}_t^{att} \ , \\
    where~\mathbf{F}_t^{att} &=& \sigma(\mathtt{Conv}(\mathtt{Cat}(\mathbf{R}_t^{middle}, \mathbf{M}_t))) \ ,
\end{aligned}
\end{equation}
where $\sigma$ denotes sigmoid activation; $\mathtt{Cat}$ denotes concatenation operation. Then, the attention-augmented feature $\mathbf{F}_t^{ph}$ will be passed to the final multi-characteristics fusion.

\subsection{Multi-Characteristics fusion}
\label{sec:mcf}
In this section, we present a dedicated feature fusion module to aggregate physical/spatio/temporal characteristic features. 
Specifically, as shown in Figure~\ref{fig:framework}, first, the physical features $\mathbf{F}_t^{ph}$ and spatio features $\mathbf{F}_t^{sp}$ are concatenated. 
Then we introduce three Temporal-aware Attention Blocks (TAB) to further fuse the temporal features $\mathbf{F}_t^{te}$.
It is worth noting that due to the different ways in which temporal information and spatial information are encoded, we consider using temporal features for filtering the important channels of hybrid features instead of direct concatenation.
After each TAB, the generated features will also be added with the upsampled features $\{\mathbf{D}_i\}_{i=1}^{3}$ from the shadow removal decoder in the physical branch to obtain richer scale features. The output of TABs can be formulized as:
\begin{equation}
\label{Equ:TAB}
\left\{
\begin{aligned}
\mathbf{B}^{out}_{1} &=& &\mathrm{TAB}_{1}(\mathtt{Cat}(\mathbf{F}_t^{ph}, \mathbf{F}_t^{sp}), \mathbf{F}_t^{te}), \\
\mathbf{B}^{out}_{i} &=& &\mathrm{TAB}_{i}(\mathbf{B}^{out}_{i-1} + \mathbf{D}_{i-1}, \mathbf{F}_t^{te}), i = 2, 3,
\end{aligned}
\right.
\end{equation}
where $\mathrm{TAB}_{i}$ denotes each TAB block; $\mathbf{B}^{out}_{i}$ denotes the output of each TAB block.
Finally, the last hybrid features will add to the original input $\mathbf{I}_{t}$ to get the final shadow removal output $\mathbf{R}_{t}^{final}$ as follows:
\begin{equation}\label{Equ:TAB2}  
    \mathbf{R}_t^{final} = \mathbf{I}_t + \mathtt{Conv}(\mathbf{B}^{out}_{3} + \mathbf{D}_{3}),
\end{equation}
Note that the fusion module operates on the original input image resolution.
$\mathbf{F}_t^{ph}$ and $\mathbf{F}_t^{te}$ should be upsampled to $W \times H$ before being fed into the fusion module. 
Meanwhile, there is no downsampling operation in fusion module, thereby preserving the desired fine texture in the final output image.

\noindent
\textbf{Temporal-aware Attention Block.} TAB mainly consists of two types of channel attention blocks. Previous works~\cite{zhang2018image,zamir2021multi} demonstrate that the stacking of channel attention modules can help to achieve significant performance gains in most image restoration tasks.
In our work, we follow this progressive processing manner as \cite{zamir2021multi} and consider the extra temporal features $\mathbf{F}_t^{te}$ to make the model be aware of temporal information.

\begin{figure}[]
\centering
\includegraphics[width=0.5\textwidth]{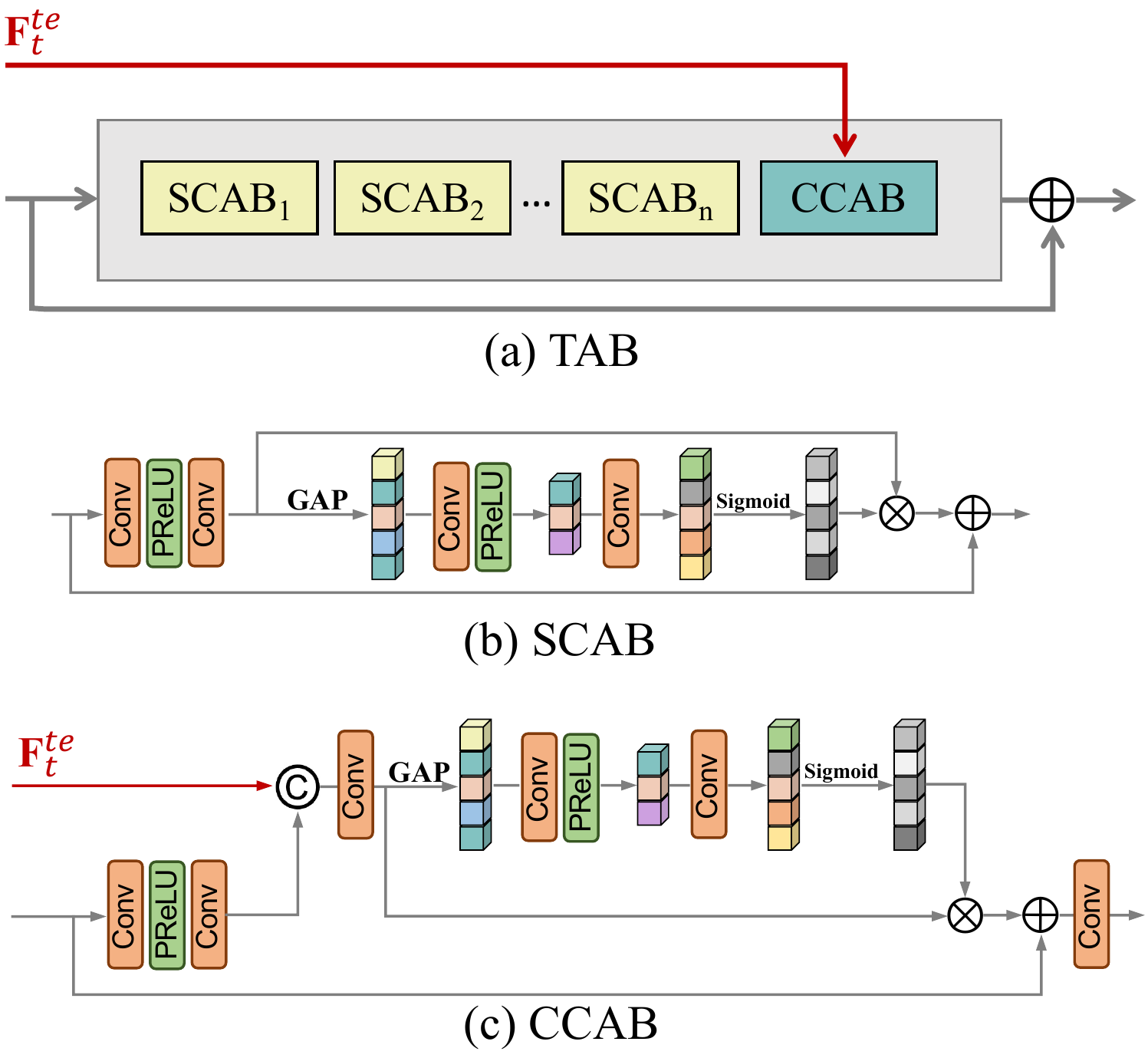}
\caption{(a) is the schematic illustration of our Temporal-aware Attention Block (TAB); (b) is one of the self channel attention block (SCAB) in (a); (c) is the cross channel attention block (CCAB) in (a). }
\label{fig:TAB}
\vspace{-2.5mm}
\end{figure}

As illustrated in Figure~\ref{fig:TAB} (a), TAB includes $k$ self channel attention blocks (SCAB) and one cross channel attention block (CCAB), while TAB adopts residual learning to make convergence easier. In our experiments, we set $k$=$8$. SCAB takes progressive hybrid features as input, and CCAB takes progressive hybrid features and temporal features $\mathbf{F}_t^{te}$ as inputs. 
Figure~\ref{fig:TAB} (b) show the details of SCAB.
It first encodes shadow removal features with two $3$$\times$$3$ convolution followed by the PReLU activation. The global information is extracted via a global average pooling operation (GAP), then fed into two $1$$\times$$1$ convolutions followed by a sigmoid activation to achieve the channel attention maps. Finally, the augmented features re-calibrated by these channel attention maps will be passed to the next block.
Figure~\ref{fig:TAB} (c) show the details of CCAB.
It first concatenates the $\mathbf{F}_t^{te}$ to the main branch hybrid features. Next, a simple $1$$\times$$1$ convolution is used to lower the dimension of these hybrid features, then the rest process is same as SCAB. 
In this way, $\mathbf{O}_{t, t+1}$ is encoded into the shadow removal features and controls the expression of temporal information.

\subsection{Loss Function}
\label{sec:loss}
There are three types of loss in our network: (1) regression loss ($\mathcal{L}_{reg}$) for over-exposure parameters $\{[\mathrm{w}_t^n, \mathrm{b}_t^n]\}_{n=1}^N$ in AEEM; (2) segmentation loss ($\mathcal{L}_{seg}$) for shadow detection mask $\mathbf{F}_{t}^{msk}$ in MSAM; (3) restoration loss ($\mathcal{L}_{res}$) for shadow removal predictions $\mathbf{R}_{t}^{middle}$ in MSAM and also $\mathbf{R}_{t}^{final}$ in the fusion module. 
The total loss $\mathcal{L}_{total}$ can be written as follows:
\begin{equation}\label{Equ:TotalLoss}  
    \mathcal{L}_{total} = \alpha \mathcal{L}_{reg} + \beta \mathcal{L}_{seg} + \gamma \mathcal{L}_{res},
\end{equation}
where $[\alpha, \beta, \gamma]$ is the trade-off weight. In our experiments, we set them all to 1.

Regression loss $\mathcal{L}_{reg}$ in Equ~\ref{Equ:TotalLoss} can be further formulated as:
\begin{equation}\label{Equ:RegLoss}  
    \mathcal{L}_{reg} = \sum_{n=1}^{N}{\big{(}\mathrm{\Phi}_{char}(\mathbf{w}_{t}^{n}, \mathbf{\hat{w}}_{t}^{n}) + \mathrm{\Phi}_{char}(\mathbf{b}_{t}^{n}, \mathbf{\hat{b}}_{t}^{n})\big{)}},
\end{equation}
where $\mathrm{\Phi}_{char}$ is Charbonnier loss~\cite{charbonnier1994two} (similar as L1 loss and smoother than L1 loss around zero point); 
$N$ represents the patch number for AEEM (in our experiments, we set $N$=$4$); $\mathbf{w}_{t}^{n}$ and $\mathbf{b}_{t}^{n}$ are $n$-th patch estimation of $[\mathrm{w}, \mathrm{b}]$ with AEEM for the $t$-th frame; $\mathbf{\hat{w}}_{t}^{n}$ and $\mathbf{\hat{b}}_{t}^{n}$ are the corresponding ground-truth. 
The generation details of $\mathbf{\hat{w}}_{t}^{n}$ and $\mathbf{\hat{b}}_{t}^{n}$ are described in Section~\ref{sec:details}.

Segmentation loss $\mathcal{L}_{seg}$ in Equ~\ref{Equ:TotalLoss} can be formulated  as:
\begin{equation}\label{Equ:SegLoss}  
    \mathcal{L}_{seg} = \mathrm{\Phi}_{bce}(\mathbf{M}_{t}, \mathbf{\hat{M}}_t),
\end{equation}
where $\mathrm{\Phi}_{bce}$ is binary cross-entropy loss~\cite{rubinstein2004cross}; $\mathbf{M}_{t}$ represents shadow mask prediction for the $t$-th frame; $\mathbf{\hat{M}}_{t}$ is the corresponding ground-truth. 

Restoration loss $\mathcal{L}_{res}$ in Equ~\ref{Equ:TotalLoss} can be further written as:
\begin{equation}\label{Equ:ResLoss}  
    \mathcal{L}_{res} = \mathrm{\Phi}_{char}(\mathbf{R}_{t}^{middle}, \mathbf{\hat{R}}_t) + \mathrm{\Phi}_{char}(\mathbf{R}_{t}^{final}, \mathbf{\hat{R}}_t),
\end{equation}
where $\mathrm{\Phi}_{char}$ is Charbonnier loss; $\mathbf{R}_{t}^{middle}$ and $\mathbf{R}_{t}^{final}$ denote coarse and refined shadow removal results for the $t$-th frame, respectively; $\mathbf{\hat{R}}_{t}$ is the corresponding ground-truth which denotes the shadow free image.

\subsection{Model adaptation in real world scenes}
\label{sec:S2R}
In the previous section, we introduce the PSTNet for video shadow removal. Due to the lack of available video shadow pairs in real world scene, in this paper, we build a synthetic dataset SVSRD-85 (details in Section~\ref{sec:dataset}) and train models with the synthetic video shadow pairs. However, there is large domain gap between synthetic scenes and real world scenes. When we trivially apply the models that trained in synthetic scenes to the real world scenes, the models tend to fail. In this section, we propose a lightweight synthetic-to-real strategy, termed S2R, to adapt the synthetic-driven models to the real world scenes without retraining.

\begin{figure}[t]
\centering
\includegraphics[width=0.45\textwidth]{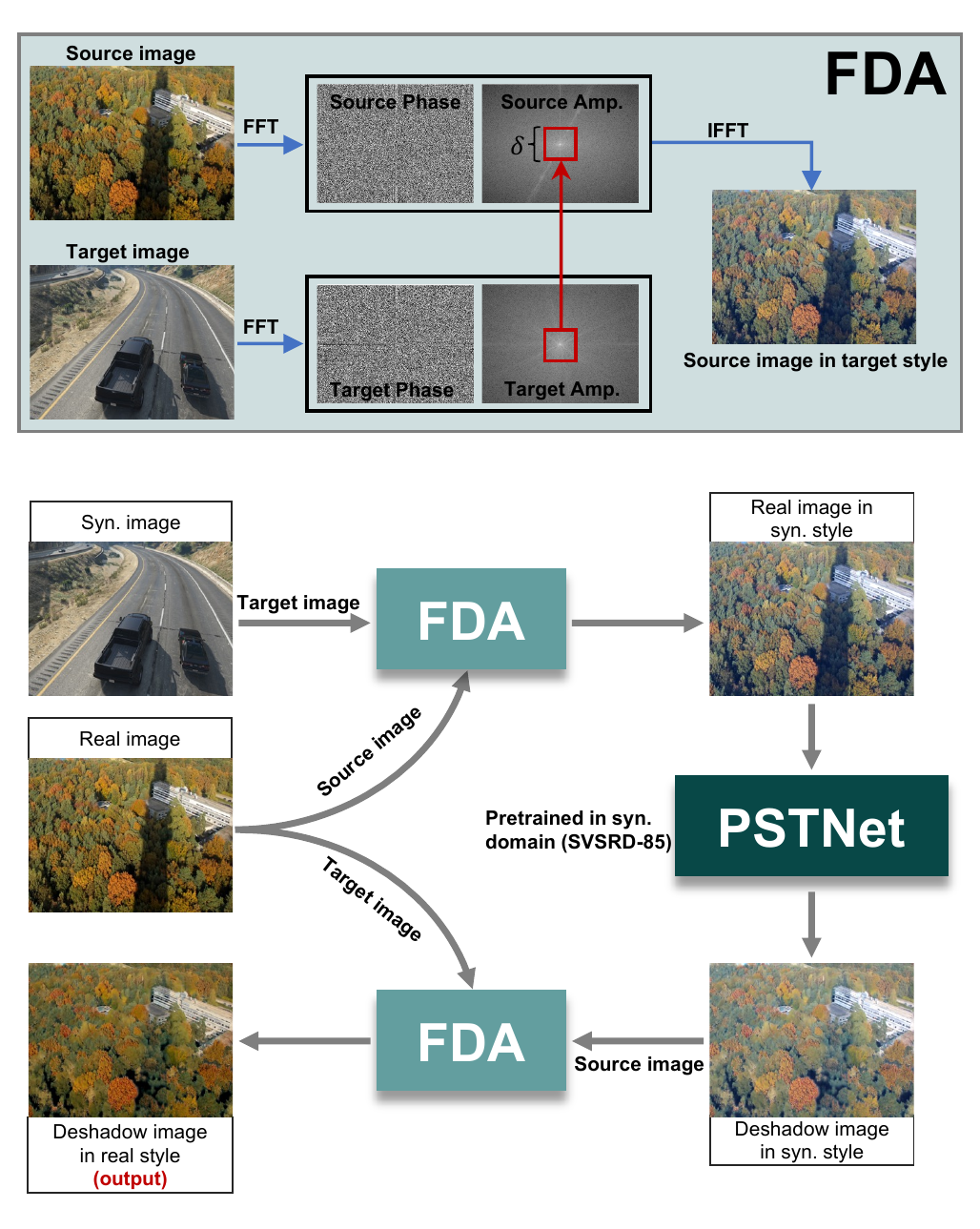}
\caption{Illustration of our proposed S2R strategy. In above figure, `PSTNet' denotes our proposed PSTNet model that pretrained in synthetic dataset (SVSRD-85). `FDA' denotes the Fourier Domain Adaptation operation~\cite{yang2020fda}.}
\label{fig:FDA}
\vspace{-2.5mm}
\end{figure}

Figure~\ref{fig:FDA} (lower) illustrate the process of S2R. First, each frame from a real world video is adapted to the synthetic domain by Fourier Domain Adaptation (FDA)~\cite{yang2020fda}. Second, a pretrained synthetic-driven video shadow removal model (e.g. PSTNet) is used to remove shadow in synthetic domain. Finally, the deshadow image in synthetic domain is adapted to real domain by FDA again, further produce the final shadow removal result. The total process can be formulized as:
\begin{equation}\label{Equ:S2R}
    \mathbf{R}_{rea} = \mathrm{FDA}(\mathrm{PSTNet}(\mathrm{FDA}(\mathbf{I}_{rea}, \mathbf{I}_{syn})), \mathbf{I}_{rea}),
\end{equation}
where $\mathbf{I}_{rea}$ and $\mathbf{I}_{syn}$ denote the input real image and synthetic image, respectively; $\mathbf{R}_{rea}$ denotes the shadow removed result of $\mathbf{I}_{rea}$; $\mathrm{FDA}(\mathtt{source}, \mathtt{target})$ denotes the FDA operation with inputs of source and target images. S2R requires only Fourier-based style transformation for the input video. With S2R, real world video can be produced trivially by synthetic-driven video shadow removal model without retraining.

\noindent
\textbf{Fourier Domain Adaptation (FDA).}
FDA~\cite{yang2020fda} is first introduced for unsupervised domain adaptation, whereby the discrepancy between the source and target distributions is reduced by swapping the low-frequency spectrum of one with the other. FDA does not require any training to perform the domain alignment, just a simple Fourier Transform and its inverse. Here, we employ FDA to reduce the domain gap between test data (from real world) and pretrained data (from synthetic scenes). Figure~\ref{fig:FDA} (upper) illustrate the process of FDA. FDA aims to adapt the source image to the target image style. First, the RGB-based source image and target image are transformed to amplitude and phase components via Fourier transform~\cite{frigo1998fftw}, respectively. Second, the low frequency part (controlled by a hyper parameter $\delta$) of the amplitude of the source image is replaced by the counterpart of the amplitude of the target image. Finally, inverse Fourier Transform~\cite{frigo1998fftw} reassemble the source phase and the new amplitude to obtain the `source image in target style'. The details of FDA can be found in \cite{yang2020fda}.

\noindent
\textbf{Choice of reference synthetic image}
Considering each frame in real world video, in S2R framework, we need to select a reference synthetic image for FDA transformation. In order to make the image transition natural, we tend to choose reference image with small gap in color space. In this paper, we use color histogram~\cite{novak1992anatomy} to filter the reference image. Specifically, we extract the first frame of each video in SVSRD-85 and compute the color histograms of them, termed $\{h_i\}_{i=1}^{s}$, where $i$ denotes the video index, $h_i$ denotes i-th color histogram and $s$ denotes number of videos in SVSRD-85. To a real world video to be processed, we also compute the color histogram of the first frame, termed $\tilde{h}$, and then computer the similarity scores between $\tilde{h}$ and $\{h_i\}_{i=1}^{s}$ with the following formulation: $score_i = |\tilde{h}-h_i|$. Finally, We choose the candidate image corresponding to the minimum value of similarity score as the final reference synthetic image as follows:
\begin{equation}\label{Equ:score}
    \hat{i} = \mathop{\mathrm{argmin}}\limits_{i\in\{1,...s\}}|\tilde{h}-h_i|,
\end{equation}
where $\hat{i}$ is the final selected reference synthetic image.

\section{Experiments}
\label{sec:experiments}
\subsection{Fundamental Settings}
\label{sec:details}
\noindent
\textbf{Evaluation measures.}
Following previous works~\cite{wang2018stacked,guo2012paired,qu2017deshadownet,le2021physics,fu2021auto}, we utilize the root mean square error (RMSE) in LAB color space between the predicted shadow removal result and the ground-truth image to evaluate different shadow removal methods. 

\noindent
\textbf{Comparative Methods.}
We make comparison against nine state-of-the-art methods for relevant tasks, including Guo \textit{et al.}~\cite{guo2012paired}, Gong \textit{et al.}~\cite{gong2014interactive}, DSC~\cite{hu2019direction}, SID~\cite{le2021physics}, DHAN~\cite{cun2020towards}, and Expo~\cite{fu2021auto} for single image shadow removal; MPRNet~\cite{zamir2021multi} for single image restoration; EDVR~\cite{wang2019edvr} for video restoration; and BasicVSR~\cite{chan2021basicvsr} for video super-resolution. 
We utilize their public codes, and re-train these methods on the SVSRD-85 training set to produce their best results for a fair comparison.

\noindent
\textbf{Implementation Details}
Our PSTNet is end-to-end trainable and requires no pre-training. 
The networks are trained on $256$$\times$$256$ patches on two NVIDIA GTX 2080Ti by using an Adam optimizer with a batch size of 6, 200 epochs, and an initial learning rate of $2$$\times$$10^{-4}$. 
The learning rate is then steadily decreased to $1$$\times$$10^{-6}$ using a cosine annealing strategy~\cite{loshchilov2016sgdr}. 
For data augmentation, horizontal and vertical flips are randomly applied. 
Note that our SVSRD-85 dataset provides the ground-truth of shadow-free image $\mathbf{\hat{R}_t}$ and shadow mask $\mathbf{\hat{M}_t}$ for each video frame.
Moreover, like SID~\cite{le2021physics}, we generate the ground-truth $\{[\mathrm{\hat{w}}_t^n, \mathrm{\hat{b}}_t^n]\}_{n=1}^N$ for AEEM, using a least squares method regression~\cite{chatterjee1986influential} via $\mathbf{\hat{I}_t}$, $\mathbf{\hat{R}_t}$ and $\mathbf{\hat{M}_t}$.

\begin{figure*}[]
\centering
\includegraphics[width=1\textwidth]{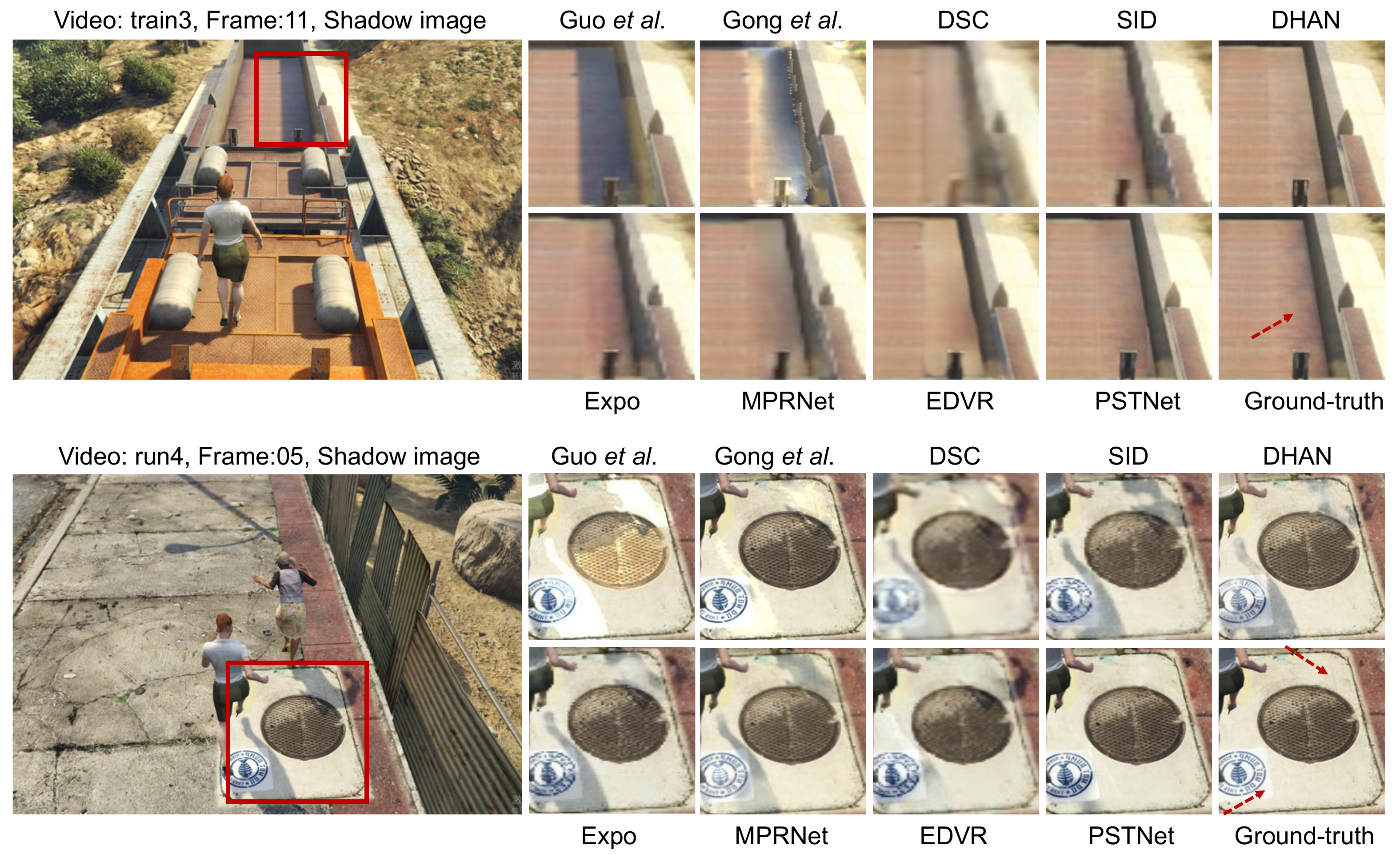}
\vskip -5pt
\caption{Qualitative comparison between our methods and other shadow removal methods on our SVSRD-85 dataset.}
\label{fig:visual_SVSRD-85}
\vspace{-2.5mm}
\end{figure*}
\subsection{Comparison with the State-of-the-arts}
\begin{table}[t]
\begin{center}
  \caption{Comparing our network (PSTNet) against the state-of-the-art methods on our proposed SVSRD-85 dataset.}
  \vskip -5pt
  \label{table:state-of-the-art}
  \resizebox{0.46\textwidth}{!}{%
    \begin{tabular}{c|c|c|c|c}
        \toprule[1pt]
        Method~~ $\backslash$ ~~RMSE & Year & \textbf{Shadow} & \textbf{Non-Shadow} & \textbf{All} \\
        \specialrule{0em}{1pt}{1pt}
        \hline
        \specialrule{0em}{1pt}{1pt}
        Input Image & - & 42.08 & 6.26 & 11.82 \\
        \specialrule{0em}{1pt}{1pt}
        \hline
        \specialrule{0em}{1pt}{1pt}
        Guo \textit{et al.}~\cite{guo2012paired}& 2013 & 25.54 & 13.32 & 15.18 \\
        Gong \textit{et al.}~\cite{gong2014interactive} & 2014 & 18.90 & 6.91 & 8.67 \\
        DSC~\cite{hu2019direction} & 2019 & 18.28 & 9.91 & 11.19 \\
        SID~\cite{le2021physics} & 2019 & 15.69 & 8.69 & 9.49 \\
        DHAN~\cite{cun2020towards} & 2020 & 14.97 & 9.87 & 10.49 \\
        Expo~\cite{fu2021auto} & 2021 & 16.34 & 8.66 & 9.68 \\
        MPRNet~\cite{zamir2021multi} & 2021 & 16.57 & 7.73 & 8.99 \\
        \specialrule{0em}{1pt}{1pt}
        \hline
        \specialrule{0em}{1pt}{1pt}
        EDVR~\cite{wang2019edvr} & 2019 & 16.50 & 9.47 & 10.30 \\
        BasicVSR~\cite{chan2021basicvsr} &2021 & 17.78 & 9.91 & 10.93 \\
        \specialrule{0em}{1pt}{1pt}
        \hline
        \specialrule{0em}{1pt}{1pt}
        \textbf{PSTNet(ours)} & - & \textbf{12.77} & \textbf{6.18} & \textbf{6.93}\\
        \specialrule{0em}{1pt}{1pt}
        \bottomrule[1pt]
    \end{tabular}
    }
    \vspace{-5mm}
  \end{center}
\end{table}

Table~\ref{table:state-of-the-art} lists RMSE scores of our PSTNet and compared methods at shadow pixels, non-shadow pixels, and all pixels of the whole video frames from our SVSRD-85 dataset.
And the first row shows the RMSE values of the input shadow videos without any shadow removal operation.
Among all the methods, our PSTNet obtains the smallest RMSE scores at shadow regions, non-shadow regions, and the whole image, which indicate that our network has better video shadow removal performance than the compared methods. 
Specifically, compared against two encoder-decoder based image shadow removal methods (i.e., DSC~\cite{hu2019direction} and DHAN~\cite{cun2020towards}), our PSTNet outperforms DSC by 30.1\%/37.6\% RMSE in shadow/non-shadow region and outperforms DHAN by 14.7\%/37.3\% in shadow/non-shadow region. 
Compared with physical-based methods SID~\cite{le2021physics} and Expo~\cite{fu2021auto}, PSTNet outperforms SID by 18.6\%/28.8\% in shadow/non-shadow region and outperforms Expo by 21.8\%/28.6\% in shadow/non-shadow region. 
PSTNet also outperforms the image restoration method MPRNet~\cite{zamir2021multi} by 22.9\%/20.0\%. 
In addition, compared with the video restoration method EDVR~\cite{wang2019edvr} and video super-resolution method BasicVSR~\cite{chan2021basicvsr}, PSTNet also outperforms EDVR by 22.6\%/34.7\% in shadow/non-shadow region and outperforms BasicVSR by 28.1\%/37.6\% in shadow/non-shadow region.

Figure~\ref{fig:visual_SVSRD-85} visually compares the shadow removal results produced by our network and other methods on the SVSRD-85 dataset. 
For the shadow on the deck of the train (first case), traditional methods (Guo \textit{et al.}~\cite{guo2012paired} and Gong \textit{et al.}~\cite{gong2014interactive}) can not distinguish this region as shadow, thereby generating the predictions that are similar as input. 
Most data-driven methods can remove the shadow of the central part to some extent, while they all tend to generate the ghost shadow over the original shadow boundary.
In comparison, our PSTNet can yield reasonable exposure estimation and smooth removal result on the shadow boundary. 
For the second case, the ground over the manhole cover is white, which makes the illustration of the shadow cast on it higher than the ground regions. 
Most image-based shadow removal methods cannot remove the persons' shadow due to being cheated by the high values of shadows. 
However, the video-based methods like EDVR~\cite{wang2019edvr} and our PSTNet take advantage of the hints of temporal information between frames, which help to distinguish shadows from lit regions. Furthermore, the AEEM in PSTNet can help generate adaptive exposure estimation for different regions, thereby preventing from generating the discontiguous prediction between adjacent regions with different textures, like EDVR.

\begin{figure*}[t]
	\centering
	\vspace*{0.5mm}
    \begin{subfigure}{0.16\textwidth} 
		\includegraphics[width=\textwidth]{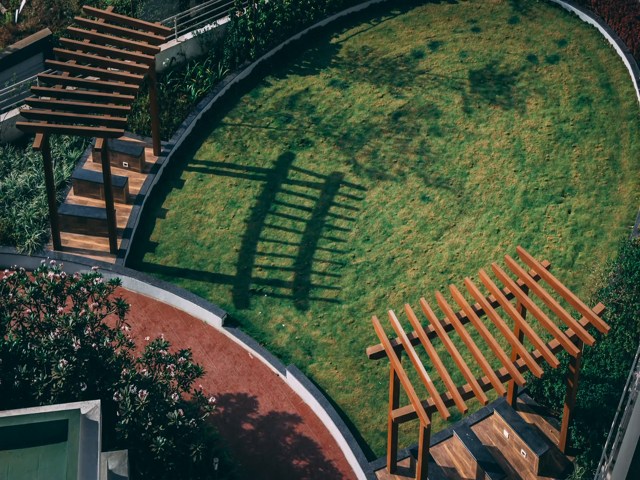}
	\end{subfigure}
	\begin{subfigure}{0.16\textwidth}
		\includegraphics[width=\textwidth]{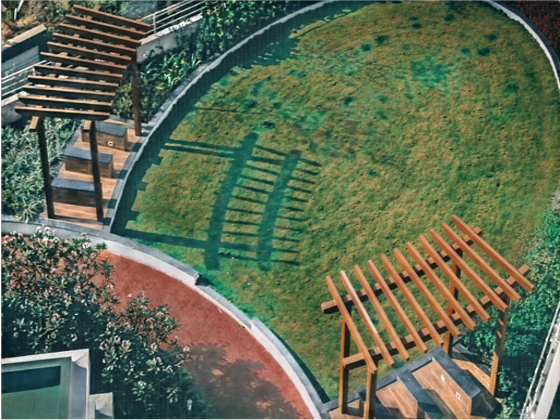}
	\end{subfigure}
	\begin{subfigure}{0.16\textwidth}
		\includegraphics[width=\textwidth]{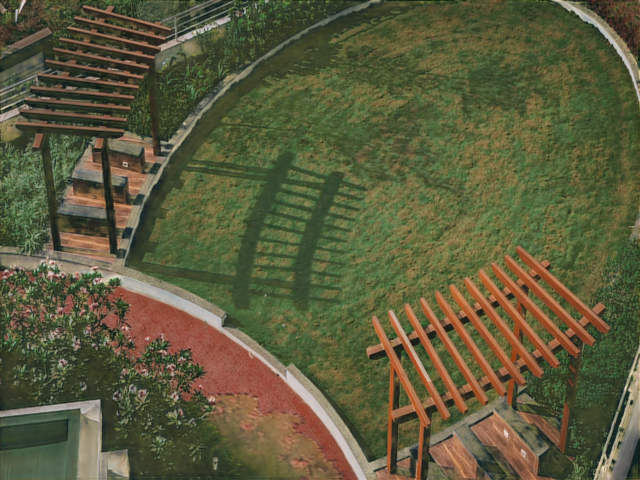}
	\end{subfigure}
	\begin{subfigure}{0.16\textwidth}
		\includegraphics[width=\textwidth]{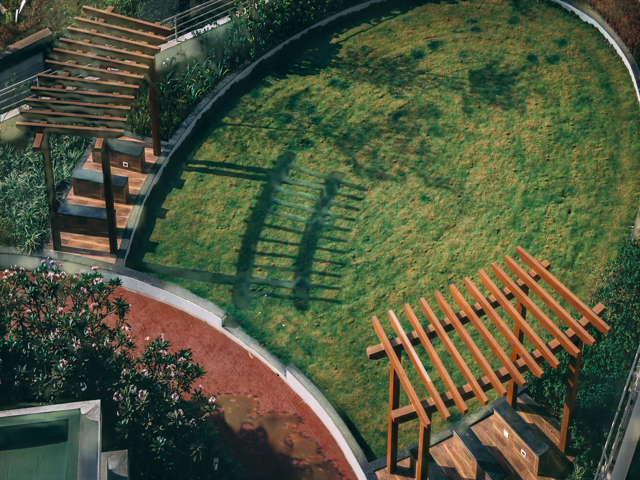}
	\end{subfigure}
	\begin{subfigure}{0.16\textwidth}
		\includegraphics[width=\textwidth]{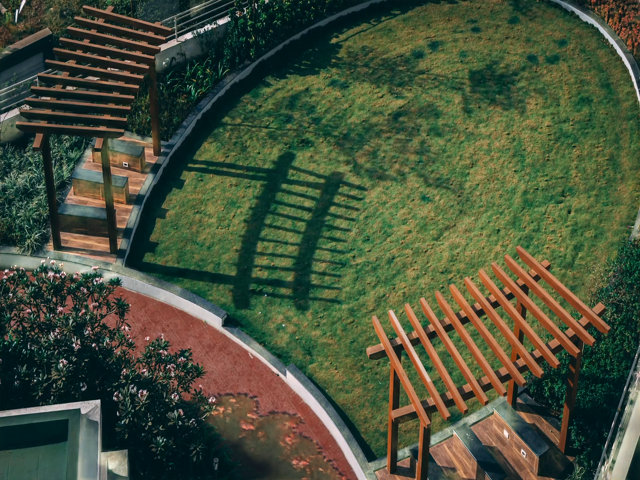}
	\end{subfigure}
	\begin{subfigure}{0.16\textwidth}
		\includegraphics[width=\textwidth]{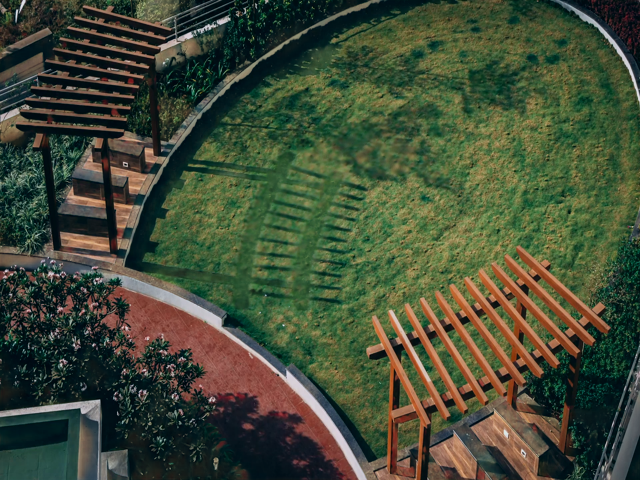}
	\end{subfigure}
	\ \\
	\vspace*{0.5mm}
    \begin{subfigure}{0.16\textwidth} 
		\includegraphics[width=\textwidth]{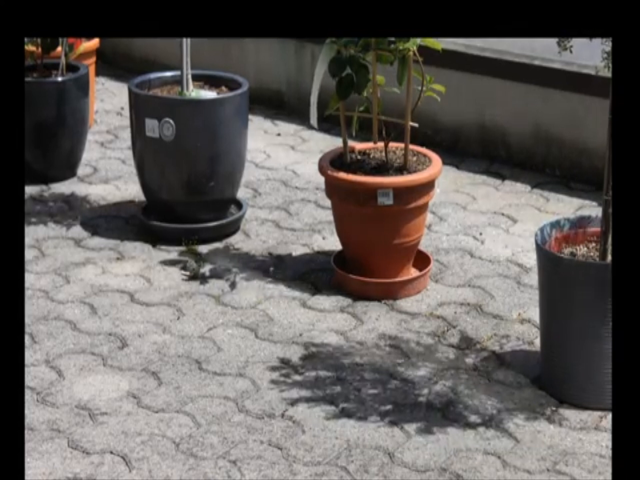}
	\end{subfigure}
	\begin{subfigure}{0.16\textwidth}
		\includegraphics[width=\textwidth]{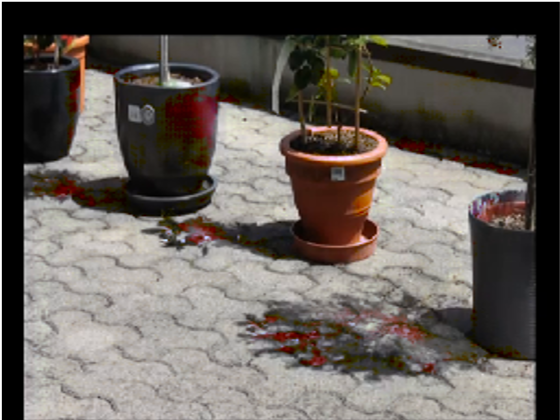}
	\end{subfigure}
	\begin{subfigure}{0.16\textwidth}
		\includegraphics[width=\textwidth]{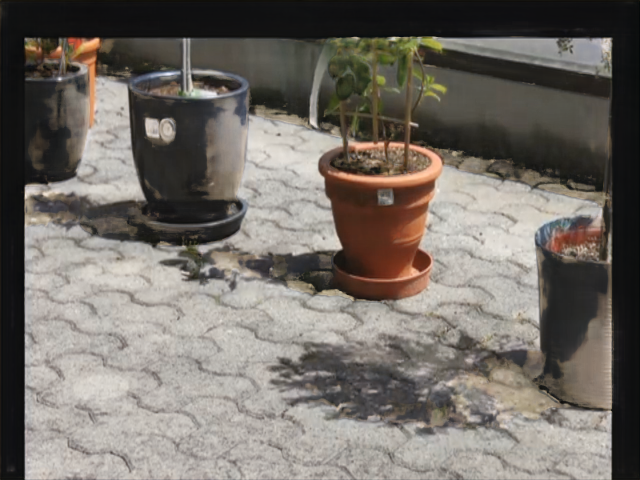}
	\end{subfigure}
	\begin{subfigure}{0.16\textwidth}
		\includegraphics[width=\textwidth]{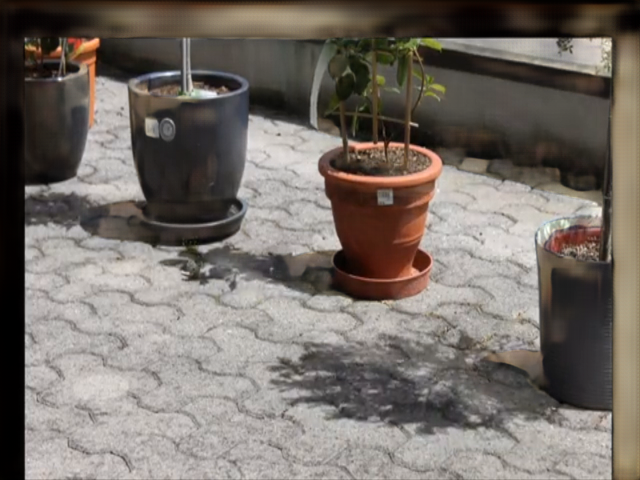}
	\end{subfigure}
	\begin{subfigure}{0.16\textwidth}
		\includegraphics[width=\textwidth]{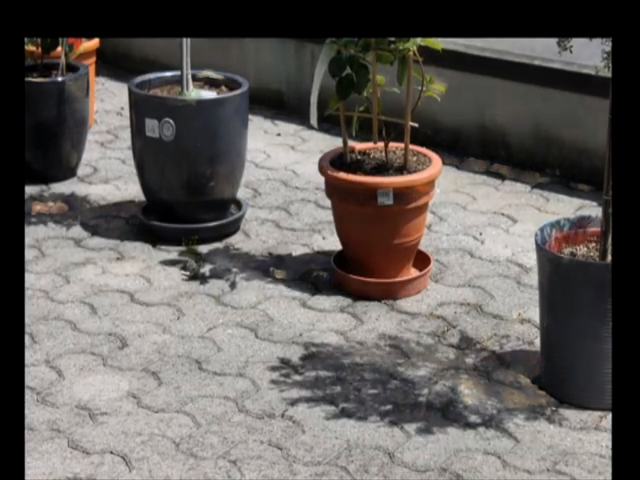}
	\end{subfigure}
	\begin{subfigure}{0.16\textwidth}
		\includegraphics[width=\textwidth]{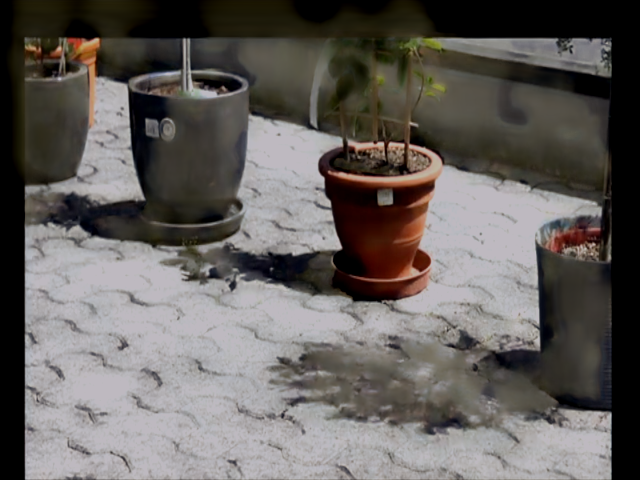}
	\end{subfigure}
	\ \\
	\vspace*{0.5mm}
    \begin{subfigure}{0.16\textwidth} 
		\includegraphics[width=\textwidth]{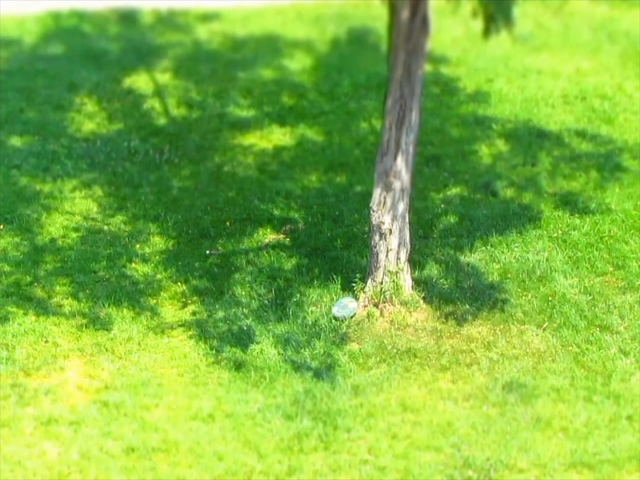}
	\end{subfigure}
	\begin{subfigure}{0.16\textwidth}
		\includegraphics[width=\textwidth]{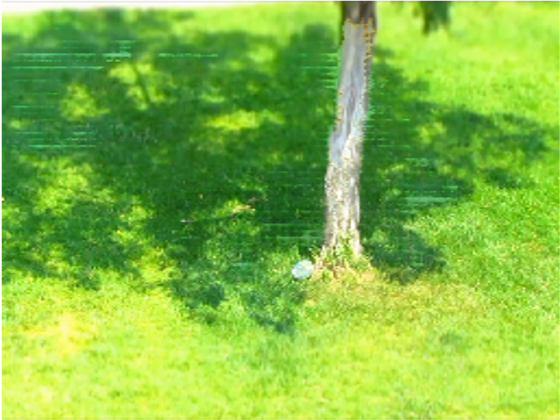}
	\end{subfigure}
	\begin{subfigure}{0.16\textwidth}
		\includegraphics[width=\textwidth]{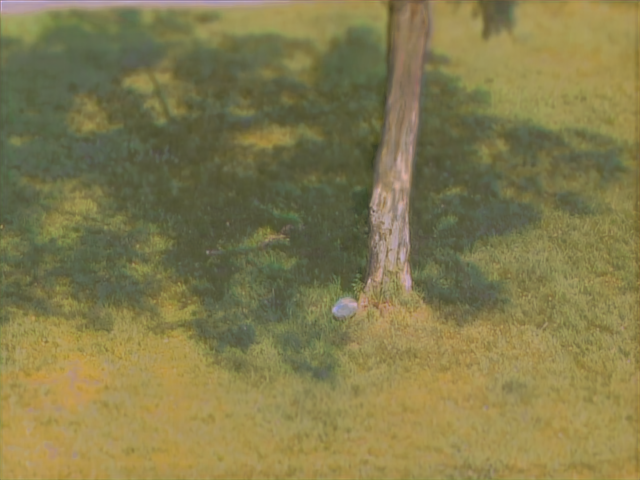}
	\end{subfigure}
	\begin{subfigure}{0.16\textwidth}
		\includegraphics[width=\textwidth]{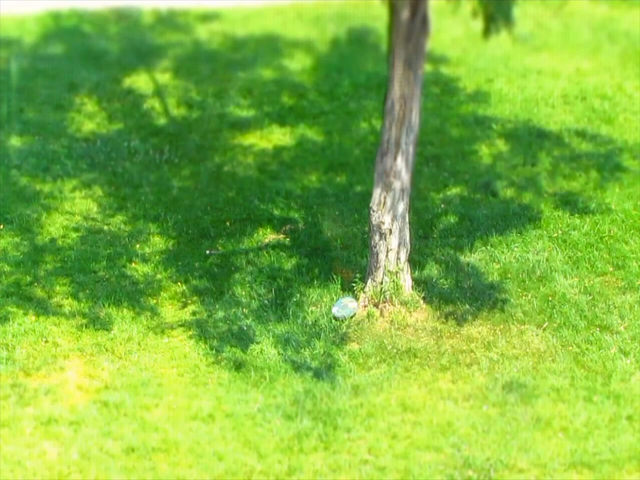}
	\end{subfigure}
	\begin{subfigure}{0.16\textwidth}
		\includegraphics[width=\textwidth]{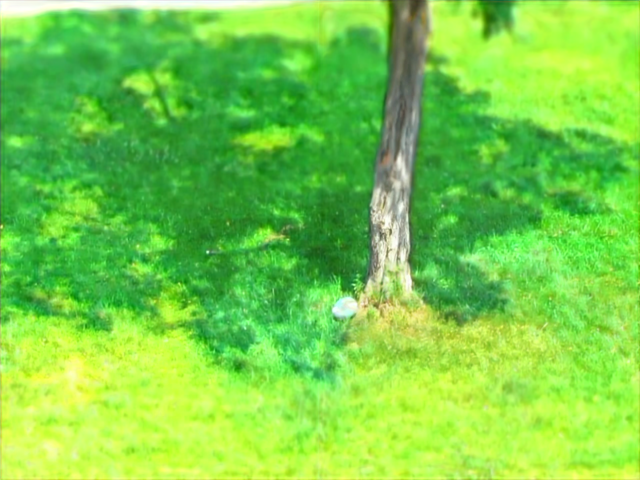}
	\end{subfigure}
	\begin{subfigure}{0.16\textwidth}
		\includegraphics[width=\textwidth]{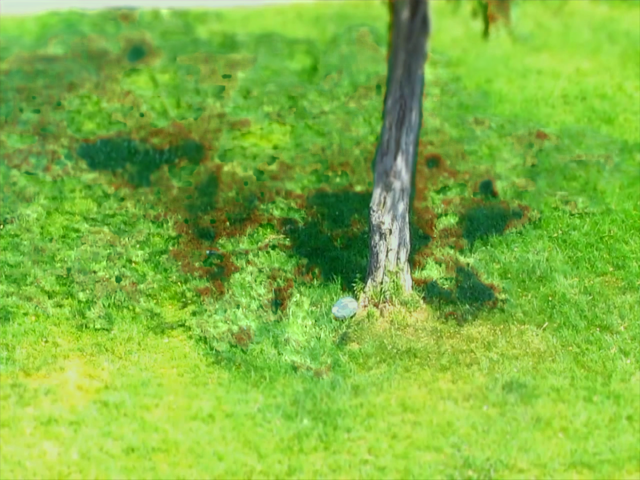}
	\end{subfigure}
	\ \\

	\vspace*{0.5mm}
	\begin{subfigure}{0.16\textwidth}
		\includegraphics[width=\textwidth]{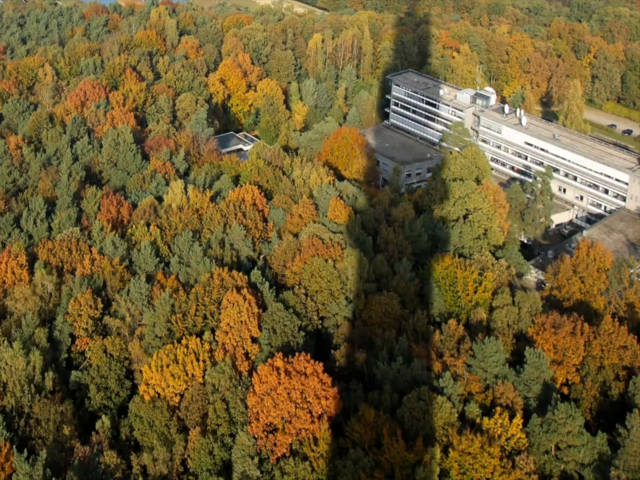}
		\vspace{-5.5mm} \caption*{{\footnotesize Input}}
        \vspace{-2mm} \caption*{\hspace*{-0.7mm}{\footnotesize images}}
	\end{subfigure}
	\begin{subfigure}{0.16\textwidth}
		\includegraphics[width=\textwidth]{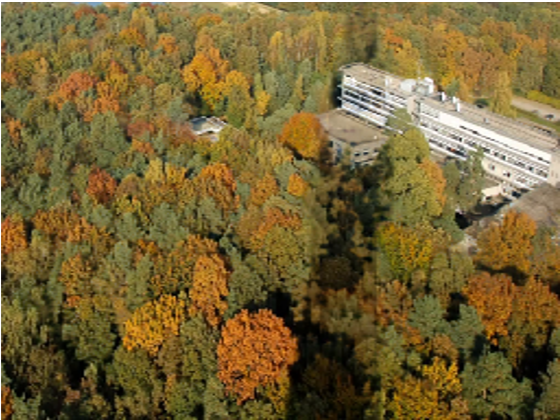}
		\vspace{-5.5mm} \caption*{{\footnotesize SID}}
        \vspace{-2mm} \caption*{\hspace*{-0.7mm}{\footnotesize ~\cite{le2021physics}}}
	\end{subfigure}
	\begin{subfigure}{0.16\textwidth}
		\includegraphics[width=\textwidth]{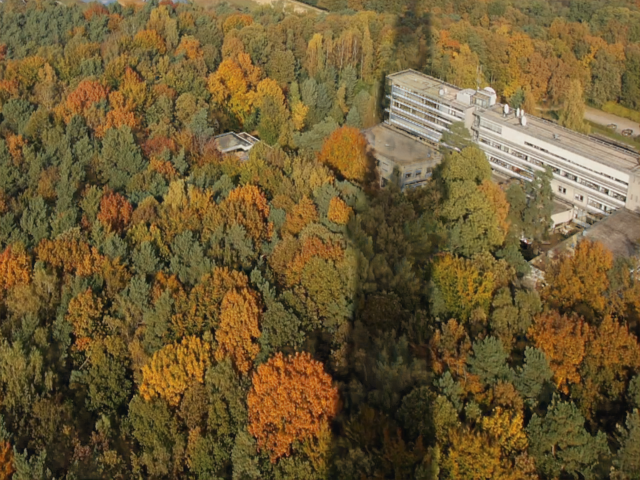}
		\vspace{-5.5mm} \caption*{{\footnotesize DHAN}}
        \vspace{-2mm} \caption*{\hspace*{-0.7mm}{\footnotesize ~\cite{cun2020towards}}}
	\end{subfigure}
	\begin{subfigure}{0.16\textwidth}
		\includegraphics[width=\textwidth]{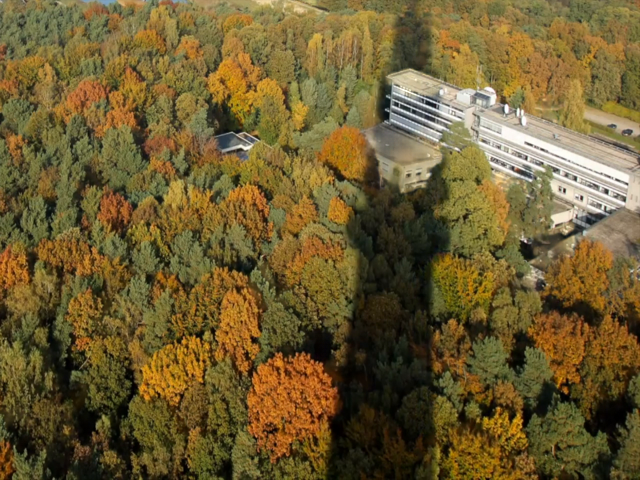}
		\vspace{-5.5mm} \caption*{{\footnotesize EDVR}}
        \vspace{-2mm} \caption*{\hspace*{-0.7mm}{\footnotesize ~\cite{wang2019edvr}}}
	\end{subfigure}
	\begin{subfigure}{0.16\textwidth}
		\includegraphics[width=\textwidth]{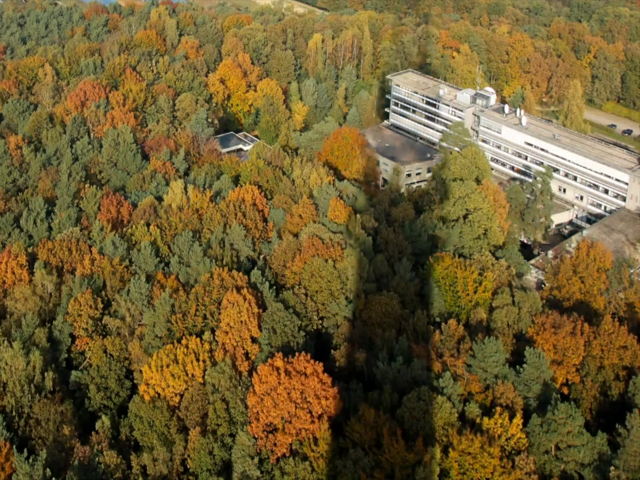}
		\vspace{-5.5mm} \caption*{{\footnotesize PSTNet}}
        \vspace{-2mm} \caption*{\hspace*{-0.7mm}{\footnotesize (ours)}}
	\end{subfigure}
	\begin{subfigure}{0.16\textwidth}
		\includegraphics[width=\textwidth]{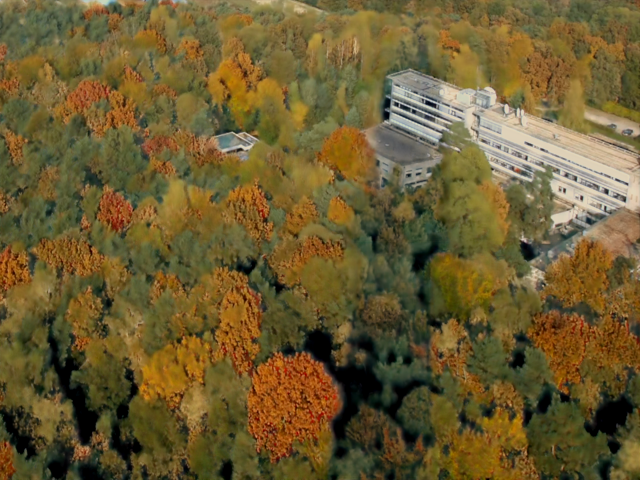}
		\vspace{-5.5mm} \caption*{{\footnotesize PSTNet + S2R}}
        \vspace{-2mm} \caption*{\hspace*{-0.7mm}{\footnotesize (ours)}}
	\end{subfigure}
    \ \\
    \vspace{-1.5mm}
	\caption{Visual comparison on the real world scenes (SBU-TimeLapse dataset). 1st column is the input frame from a specific video; 2nd-5th columns are video shadow removal results produced by our PSTNet and other comparison methods. These methods are trained on the synthetic SVSRD-85 dataset. The last column, PSTNet + FDA, is video shadow removal results produced by our PSTNet and then boosted by the `S2R' transfer strategy (mentioned in section~\ref{sec:S2R}).}
	\label{fig:comparison_TL}
    \vspace{-4.5mm}
\end{figure*}

\begin{table}[]
\begin{center}
  \caption{Quantitative results of ablation study experiments on our SVSRD-85 dataset.}
  \vskip -5pt
  \label{table:ablation}
  \resizebox{0.46\textwidth}{!}{%
    \begin{tabular}{c|c|c|c}
        \toprule[1pt]
        Method~~ $\backslash$ ~~RMSE & \textbf{Shadow} & \textbf{Non-Shadow} & \textbf{All} \\
        \specialrule{0em}{1pt}{1pt}
        \hline
        \specialrule{0em}{1pt}{1pt}
        Input Image & 42.08 & 6.26 & 11.82 \\
        \specialrule{0em}{1pt}{1pt}
        \hline
        \specialrule{0em}{1pt}{1pt}
        Physical & 16.42 & 6.91 & 8.25 \\
        Physical+Temporal & 13.92 & 6.44 & 7.36 \\
        Physical+Spatio & 13.01 & 6.38 & 7.15 \\
        \specialrule{0em}{1pt}{1pt}
        \hline
        \specialrule{0em}{1pt}{1pt}
        \textbf{Our method} & \textbf{12.77} & \textbf{6.18} & \textbf{6.93} \\
        \specialrule{0em}{1pt}{1pt}
        \hline
        \specialrule{0em}{1pt}{1pt}
        w/o AEEM & 15.69 & 6.62 & 7.95 \\
        w/o AEEM-A & 13.64 & 6.46 & 7.35 \\
         w/o MSAM & 14.01 & 6.63 & 7.67 \\
         w/o MSAM-M & 13.67 & 6.90 & 7.70 \\
         w/o FM & 14.79 & 6.88 & 7.85 \\
        
        \specialrule{0em}{1pt}{1pt}
        \bottomrule[1pt]
    \end{tabular}
    }
    \vspace{-5mm}
  \end{center}
\end{table}


\subsection{Ablation Study}
\label{sec:ablation}
We perform ablation study experiments to verify the effectiveness of three shadow characteristics and some modules of our PSTNet. 

\vspace{2mm}
\noindent
\textbf{Effectiveness of three branches.}
Here, the first baseline ``Physical'' denotes that we only exploits the physical branch of our method to remove shadows of video frames. 
The second ``Physical+Temporal'' utilizes the  physical branch, temporal branch, and feature fusion module to perform the shadow removal, 
while the third ``Physical+Spatio'' combines the physical branch, the spatio branch, and the feature fusion module to removal shadows.

Table~\ref{table:ablation} summarizes the RMSE scores of our method and three reconstructed  baseline networks on the SVSRD-85 dataset. 
From the quantitative results, we can find that
``Physical+Temporal'' has smaller RMSE scores that ``Physical'' at shadow pixels, non-shadow pixels, and all pixels of the whole video frames, which demonstrates that the temporal branch helps our method to remove shadows from video frames.
Moreover, the smaller RMSE scores of ``Physical+Spatio'' over ``Physical'' indicates that considering the spatio branch in our method incurs a better video shadow removal performance. 
More importantly, combining the three branches in our method has the best RMSE performance. 


\vspace{2mm}
\noindent
\textbf{Effectiveness of AEEM.} \  
We construct a baseline (``w/o AEEM'') by removing the adaptive exposure estimation network from our PSTNet, which means that the input shadow video frame is directly fed into the subsequent encoder for feature extraction. 
Apparently, our method consistently has smaller RMSE scores at shadow regions, non-shadow regions, and the whole video frame than ``w/o AEEM'', which shows that the exposure estimation via AEEM helps our network to better removal shadow pixels from video frames. In addition, we also construct a baseline (``w/o AEEM-A'') by replacing the adaptive exposure estimation with a fixed exposure estimation, just like SID~\cite{le2021physics}. With this setting, the shadow performance reduces from 12.77 to 13.64. This shows that dynamic adjustment of parameter estimation is necessary in some complex cases.

\vspace{2mm}
\noindent
\textbf{Effectiveness of MSAM.} \ 
We further construct a baseline (``w/o MSAM'') by removing mask-guided supervised attention module from PSTNet. Hence, ``w/o MSAM'' directly pass the decoder output of the physical branch to the subsequent fusion module without any mask supervision enhancement operation.
According to Table~\ref{table:ablation}, we can find that removing MSAM from our PSTNet degrade its video shadow removal performance due to the superior RMSE results of our PSTNet over ``w/o MSAM''. In addition, when we just remove the shadow mask prediction branch in MSAM (``w/o MSAM-M''), shadow performance reduces from 12.77 to 13.67 and the non-shadow performance reduces from. 6.18 to 6.90. This indicates that joint learning of shadow detection and removal tasks helps to improve the performance of shadow removal.

\vspace{2mm}
\noindent
\textbf{Effectiveness of multi-characteristics fusion module.} \ 
Lastly, a baseline (``w/o FM'') is reconstructed by replacing the multi-characteristics fusion module of PSTNet with a concatenation operation to assemble features form three branches. 
Apparently, our method outperforms ``w/o F.M.'' in terms of RMSE scores at shadow, non-shadow, and all pixels of the whole video frame.
It means that combining features from three branches via our multi-characteristics fusion module enables our method to reach a better video shadow removal result.

\subsection{Video shadow removal in real world scenes.}
In section~\ref{sec:S2R}, we introduce a lightweight model adaptation strategy `S2R' to adapt the well trained synthetic-driven video shadow removal model to real world scenes. We perform the generalisation experiments on the real world videos from SBU-Timelapse~\cite{le2021physics} to evaluate the effectiveness of S2R. In our experiments, we set the hyper parameters $\delta$ (mentioned in section~\ref{sec:S2R}) to 0.01. Since accurate shadow removal ground-truths in SBU-Timelapse are not available, we only perform visual comparisons on this dataset.

Figure~\ref{fig:comparison_TL} visually compares the video shadow removal results produced by our PSTNet (5-th column) and other methods (2-4 columns). The last column (6-th column) denotes the boosted PSTNet with S2R strategy. All models are trained with SVSRD-85 and the reference images in S2R are also from SVSRD-85. It can be found that S2R allows the model to adapt better to the real world scenes. For examples, in 3rd row, due to the overall bright color, the shadows of the trees are not recognized in vanilla PSTNet. However, in S2R boosted version, most of the shadows of the trees are reduced to the color of the background. In 4th row, vanilla PSTNet and other methods can only remove part of the shadows of tower, while S2R boosted version can remove almost all shadows.

\section{Limitations}
\begin{figure}[]
\centering
\includegraphics[width=0.5\textwidth]{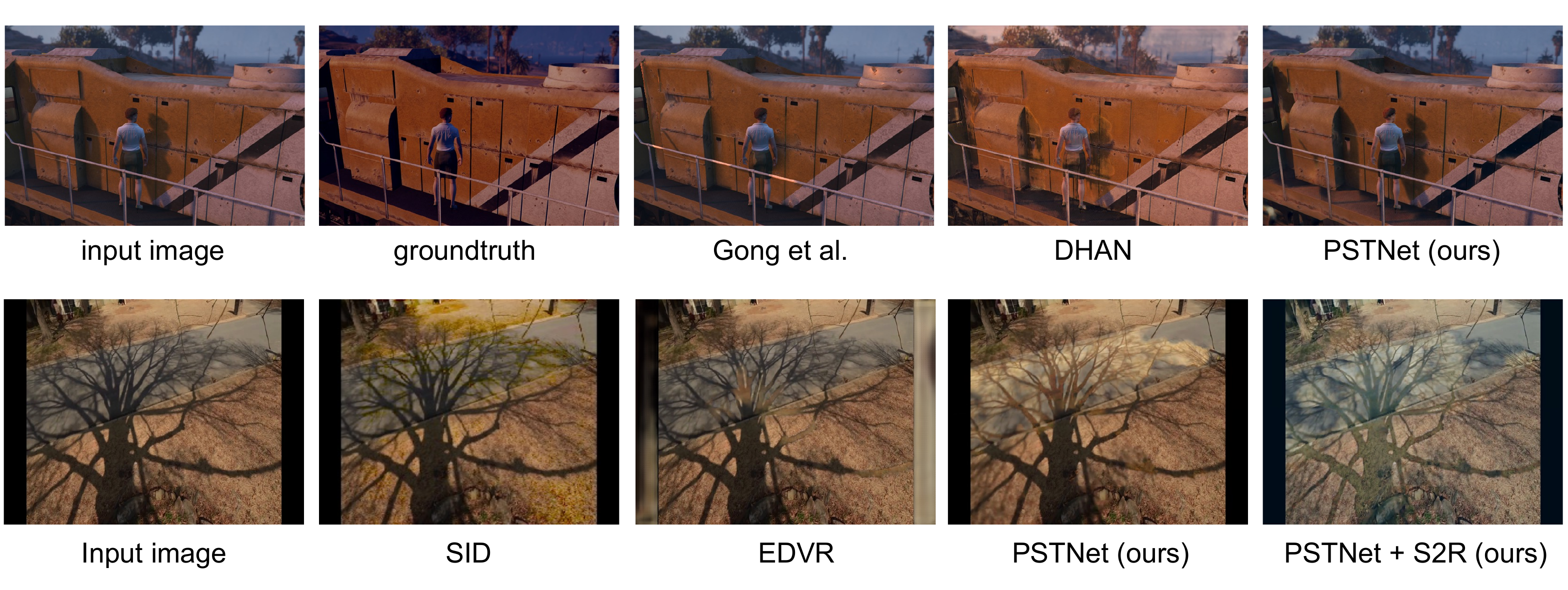}
\vskip -3pt
\caption{Some failure cases of our method.}
\label{fig:FC}
\vspace{-2.5mm}
\end{figure}
Figure~\ref{fig:FC} shows some failure cases of our method in synthetic (1st row) and real world (2nd row) scenes, respectively. From the 1st row, we can find that our method PSTNet has failed in the scenes with low illumination contrast. The same happens with DHAN~\cite{cun2020towards}, which is also based on deep learning. However, the traditional method Gong et al.~\cite{gong2014interactive} is instead very good at removing the shadows from people. From the 2nd row, we can find that there is a huge gap between synthetic and real world scenes. Despite the use of some domain adaptation approach (e.g. S2R), the shadow removal results still have large deviations in real world scenes.

\section{Conclusion}
\label{sec:conclusion}
In this paper, we have proposed a novel video shadow removal method, termed PSTNet, by considering three shadow related characteristics: physical, spatio, and temporal.
We present an improved physical characteristic to represent the impact of complex lighting and texture conditions on image shadow. 
Adaptive exposure estimation module together with mask-guided supervised attention module enable us to get reliable physical features. 
The multi-characteristic fusion module effectively combines the three kinds of features. 
Due to the lack of existing video shadow removal dataset for training models, we also collect a synthetic dataset of 85 paired videos, termed SVSRD-85. 
Extensive experiments demonstrate that our PSTNet performs best on SVSRD-85 dataset. 
Moreover, we propose a model transformation strategy to make our synthetic-driven model effective in real world scenes.
In future, we plan to further study how to overcome the domain bias between synthetic scenes and real scenes.

\bibliographystyle{IEEEtran}
\bibliography{sample-base}

\begin{thebibliography}{10}
\providecommand{\url}[1]{#1}
\csname url@samestyle\endcsname
\providecommand{\newblock}{\relax}
\providecommand{\bibinfo}[2]{#2}
\providecommand{\BIBentrySTDinterwordspacing}{\spaceskip=0pt\relax}
\providecommand{\BIBentryALTinterwordstretchfactor}{4}
\providecommand{\BIBentryALTinterwordspacing}{\spaceskip=\fontdimen2\font plus
\BIBentryALTinterwordstretchfactor\fontdimen3\font minus
  \fontdimen4\font\relax}
\providecommand{\BIBforeignlanguage}[2]{{%
\expandafter\ifx\csname l@#1\endcsname\relax
\typeout{** WARNING: IEEEtran.bst: No hyphenation pattern has been}%
\typeout{** loaded for the language `#1'. Using the pattern for}%
\typeout{** the default language instead.}%
\else
\language=\csname l@#1\endcsname
\fi
#2}}
\providecommand{\BIBdecl}{\relax}
\BIBdecl

\bibitem{okabe2009attached}
T.~Okabe, I.~Sato, and Y.~Sato, ``Attached shadow coding: Estimating surface
  normals from shadows under unknown reflectance and lighting conditions,'' in
  \emph{2009 IEEE 12th International Conference on Computer Vision}.\hskip 1em
  plus 0.5em minus 0.4em\relax IEEE, 2009, pp. 1693--1700.

\bibitem{karsch2011rendering}
K.~Karsch, V.~Hedau, D.~Forsyth, and D.~Hoiem, ``Rendering synthetic objects
  into legacy photographs,'' \emph{ACM Transactions on Graphics}, vol.~30,
  no.~6, pp. 157:1--157:12, 2011.

\bibitem{lalonde2009estimating}
J.-F. Lalonde, A.~Efros, and S.~Narasimhan, ``Estimating natural illumination
  from a single outdoor image,'' in \emph{ICCV}, 2009, pp. 183--190.

\bibitem{junejo2008estimating}
I.~Junejo and H.~Foroosh, ``Estimating geo-temporal location of stationary
  cameras using shadow trajectories,'' in \emph{ECCV}, 2008, pp. 318--331.

\bibitem{surkutlawar2013shadow}
S.~Surkutlawar and R.~K. Kulkarni, ``Shadow suppression using rgb and hsv color
  space in moving object detection,'' \emph{International Journal of Advanced
  Computer Science and Applications}, vol.~4, no.~1, 2013.

\bibitem{muller2019brightness}
T.~M{\"u}ller and B.~Erdn{\"u}e{\ss}, ``Brightness correction and shadow
  removal for video change detection with uavs,'' in \emph{Autonomous Systems:
  Sensors, Processing, and Security for Vehicles and Infrastructure 2019}, vol.
  11009.\hskip 1em plus 0.5em minus 0.4em\relax International Society for
  Optics and Photonics, 2019, p. 1100906.

\bibitem{su2016shadow}
N.~Su, Y.~Zhang, S.~Tian, Y.~Yan, and X.~Miao, ``Shadow detection and removal
  for occluded object information recovery in urban high-resolution
  panchromatic satellite images,'' \emph{IEEE Journal of Selected Topics in
  Applied Earth Observations and Remote Sensing}, vol.~9, no.~6, pp.
  2568--2582, 2016.

\bibitem{zhang2018improving}
W.~Zhang, X.~Zhao, J.-M. Morvan, and L.~Chen, ``Improving shadow suppression
  for illumination robust face recognition,'' \emph{IEEE transactions on
  pattern analysis and machine intelligence}, vol.~41, no.~3, pp. 611--624,
  2018.

\bibitem{gryka2015learning}
M.~Gryka, M.~Terry, and G.~J. Brostow, ``Learning to remove soft shadows,''
  \emph{ACM Transactions on Graphics (TOG)}, vol.~34, no.~5, pp. 1--15, 2015.

\bibitem{finlayson2005removal}
G.~D. Finlayson, S.~D. Hordley, C.~Lu, and M.~S. Drew, ``On the removal of
  shadows from images,'' \emph{IEEE transactions on pattern analysis and
  machine intelligence}, vol.~28, no.~1, pp. 59--68, 2005.

\bibitem{shor2008shadow}
Y.~Shor and D.~Lischinski, ``The shadow meets the mask: Pyramid-based shadow
  removal,'' in \emph{Computer Graphics Forum}, vol.~27, no.~2.\hskip 1em plus
  0.5em minus 0.4em\relax Wiley Online Library, 2008, pp. 577--586.

\bibitem{yang2012shadow}
Q.~Yang, K.-H. Tan, and N.~Ahuja, ``Shadow removal using bilateral filtering,''
  \emph{IEEE Transactions on Image processing}, vol.~21, no.~10, pp.
  4361--4368, 2012.

\bibitem{qu2017deshadownet}
L.~Qu, J.~Tian, S.~He, Y.~Tang, and R.~W. Lau, ``Deshadownet: A multi-context
  embedding deep network for shadow removal,'' in \emph{Proceedings of the IEEE
  Conference on Computer Vision and Pattern Recognition}, 2017, pp. 4067--4075.

\bibitem{wang2018stacked}
J.~Wang, X.~Li, and J.~Yang, ``Stacked conditional generative adversarial
  networks for jointly learning shadow detection and shadow removal,'' in
  \emph{Proceedings of the IEEE Conference on Computer Vision and Pattern
  Recognition}, 2018, pp. 1788--1797.

\bibitem{hu2019direction}
X.~Hu, C.-W. Fu, L.~Zhu, J.~Qin, and P.-A. Heng, ``Direction-aware spatial
  context features for shadow detection and removal,'' \emph{IEEE transactions
  on pattern analysis and machine intelligence}, vol.~42, no.~11, pp.
  2795--2808, 2019.

\bibitem{cun2020towards}
X.~Cun, C.-M. Pun, and C.~Shi, ``Towards ghost-free shadow removal via dual
  hierarchical aggregation network and shadow matting gan,'' in
  \emph{Proceedings of the AAAI Conference on Artificial Intelligence},
  vol.~34, no.~07, 2020, pp. 10\,680--10\,687.

\bibitem{le2021physics}
H.~Le and D.~Samaras, ``Physics-based shadow image decomposition for shadow
  removal,'' \emph{IEEE Transactions on Pattern Analysis \& Machine
  Intelligence}, no.~01, pp. 1--1, 2021.

\bibitem{fu2021auto}
L.~Fu, C.~Zhou, Q.~Guo, F.~Juefei-Xu, H.~Yu, W.~Feng, Y.~Liu, and S.~Wang,
  ``Auto-exposure fusion for single-image shadow removal,'' in
  \emph{Proceedings of the IEEE/CVF Conference on Computer Vision and Pattern
  Recognition}, 2021, pp. 10\,571--10\,580.

\bibitem{le2020shadow}
H.~Le and D.~Samaras, ``From shadow segmentation to shadow removal,'' in
  \emph{European Conference on Computer Vision}.\hskip 1em plus 0.5em minus
  0.4em\relax Springer, 2020, pp. 264--281.

\bibitem{jung2009efficient}
C.~R. Jung, ``Efficient background subtraction and shadow removal for
  monochromatic video sequences,'' \emph{IEEE Transactions on Multimedia},
  vol.~11, no.~3, pp. 571--577, 2009.

\bibitem{wang2009real}
Y.~Wang, ``Real-time moving vehicle detection with cast shadow removal in video
  based on conditional random field,'' \emph{IEEE transactions on circuits and
  systems for video technology}, vol.~19, no.~3, pp. 437--441, 2009.

\bibitem{nadimi2004physical}
S.~Nadimi and B.~Bhanu, ``Physical models for moving shadow and object
  detection in video,'' \emph{IEEE transactions on pattern analysis and machine
  intelligence}, vol.~26, no.~8, pp. 1079--1087, 2004.

\bibitem{chan2021basicvsr}
K.~C. Chan, X.~Wang, K.~Yu, C.~Dong, and C.~C. Loy, ``Basicvsr: The search for
  essential components in video super-resolution and beyond,'' in
  \emph{Proceedings of the IEEE/CVF Conference on Computer Vision and Pattern
  Recognition}, 2021, pp. 4947--4956.

\bibitem{isobe2020video1}
T.~Isobe, S.~Li, X.~Jia, S.~Yuan, G.~Slabaugh, C.~Xu, Y.-L. Li, S.~Wang, and
  Q.~Tian, ``Video super-resolution with temporal group attention,'' in
  \emph{Proceedings of the IEEE/CVF Conference on Computer Vision and Pattern
  Recognition}, 2020, pp. 8008--8017.

\bibitem{isobe2020video2}
T.~Isobe, X.~Jia, S.~Gu, S.~Li, S.~Wang, and Q.~Tian, ``Video super-resolution
  with recurrent structure-detail network,'' in \emph{European Conference on
  Computer Vision}.\hskip 1em plus 0.5em minus 0.4em\relax Springer, 2020, pp.
  645--660.

\bibitem{kim2017dynamic}
T.~H. Kim, S.~Nah, and K.~M. Lee, ``Dynamic video deblurring using a locally
  adaptive blur model,'' \emph{IEEE transactions on pattern analysis and
  machine intelligence}, vol.~40, no.~10, pp. 2374--2387, 2017.

\bibitem{zhang2018adversarial}
K.~Zhang, W.~Luo, Y.~Zhong, L.~Ma, W.~Liu, and H.~Li, ``Adversarial
  spatio-temporal learning for video deblurring,'' \emph{IEEE Transactions on
  Image Processing}, vol.~28, no.~1, pp. 291--301, 2018.

\bibitem{maggioni2021efficient}
M.~Maggioni, Y.~Huang, C.~Li, S.~Xiao, Z.~Fu, and F.~Song, ``Efficient
  multi-stage video denoising with recurrent spatio-temporal fusion,'' in
  \emph{Proceedings of the IEEE/CVF Conference on Computer Vision and Pattern
  Recognition}, 2021, pp. 3466--3475.

\bibitem{yue2020supervised}
H.~Yue, C.~Cao, L.~Liao, R.~Chu, and J.~Yang, ``Supervised raw video denoising
  with a benchmark dataset on dynamic scenes,'' in \emph{Proceedings of the
  IEEE/CVF Conference on Computer Vision and Pattern Recognition}, 2020, pp.
  2301--2310.

\bibitem{yue2021semi}
Z.~Yue, J.~Xie, Q.~Zhao, and D.~Meng, ``Semi-supervised video deraining with
  dynamical rain generator,'' in \emph{Proceedings of the IEEE/CVF Conference
  on Computer Vision and Pattern Recognition}, 2021, pp. 642--652.

\bibitem{kim2015video}
J.-H. Kim, J.-Y. Sim, and C.-S. Kim, ``Video deraining and desnowing using
  temporal correlation and low-rank matrix completion,'' \emph{IEEE
  Transactions on Image Processing}, vol.~24, no.~9, pp. 2658--2670, 2015.

\bibitem{wang2019edvr}
X.~Wang, K.~C. Chan, K.~Yu, C.~Dong, and C.~Change~Loy, ``Edvr: Video
  restoration with enhanced deformable convolutional networks,'' in
  \emph{Proceedings of the IEEE/CVF Conference on Computer Vision and Pattern
  Recognition Workshops}, 2019, pp. 0--0.

\bibitem{GTAV2015}
\BIBentryALTinterwordspacing
GTAV, ``Rockstar games: Policy on posting copyrighted rockstar games
  material,'' 2015. [Online]. Available: \url{http://tinyurl.com/pjfoqo5}
\BIBentrySTDinterwordspacing

\bibitem{guo2012paired}
R.~Guo, Q.~Dai, and D.~Hoiem, ``Paired regions for shadow detection and
  removal,'' \emph{IEEE transactions on pattern analysis and machine
  intelligence}, vol.~35, no.~12, pp. 2956--2967, 2012.

\bibitem{gong2014interactive}
H.~Gong and D.~Cosker, ``Interactive shadow removal and ground truth for
  variable scene categories.'' in \emph{BMVC}.\hskip 1em plus 0.5em minus
  0.4em\relax Citeseer, 2014, pp. 1--11.

\bibitem{zamir2021multi}
S.~W. Zamir, A.~Arora, S.~Khan, M.~Hayat, F.~S. Khan, M.-H. Yang, and L.~Shao,
  ``Multi-stage progressive image restoration,'' in \emph{Proceedings of the
  IEEE/CVF Conference on Computer Vision and Pattern Recognition}, 2021, pp.
  14\,821--14\,831.

\bibitem{xiao2013fast}
C.~Xiao, R.~She, D.~Xiao, and K.-L. Ma, ``Fast shadow removal using adaptive
  multi-scale illumination transfer,'' in \emph{Computer Graphics Forum},
  vol.~32, no.~8.\hskip 1em plus 0.5em minus 0.4em\relax Wiley Online Library,
  2013, pp. 207--218.

\bibitem{zhang2015shadow}
L.~Zhang, Q.~Zhang, and C.~Xiao, ``Shadow remover: Image shadow removal based
  on illumination recovering optimization,'' \emph{IEEE Transactions on Image
  Processing}, vol.~24, no.~11, pp. 4623--4636, 2015.

\bibitem{vicente2017leave}
T.~F.~Y. Vicente, M.~Hoai, and D.~Samaras, ``Leave-one-out kernel optimization
  for shadow detection and removal,'' \emph{IEEE Transactions on Pattern
  Analysis and Machine Intelligence}, vol.~40, no.~3, pp. 682--695, 2017.

\bibitem{ding2019argan}
B.~Ding, C.~Long, L.~Zhang, and C.~Xiao, ``Argan: Attentive recurrent
  generative adversarial network for shadow detection and removal,'' in
  \emph{Proceedings of the IEEE/CVF International Conference on Computer
  Vision}, 2019, pp. 10\,213--10\,222.

\bibitem{liu2022convnet}
Z.~Liu, H.~Mao, C.-Y. Wu, C.~Feichtenhofer, T.~Darrell, and S.~Xie, ``A convnet
  for the 2020s,'' \emph{arXiv preprint arXiv:2201.03545}, 2022.

\bibitem{zhu2022bijective}
Y.~Zhu, J.~Huang, X.~Fu, F.~Zhao, Q.~Sun, and Z.-J. Zha, ``Bijective mapping
  network for shadow removal,'' in \emph{Proceedings of the IEEE/CVF Conference
  on Computer Vision and Pattern Recognition}, 2022, pp. 5627--5636.

\bibitem{hu2019mask}
X.~Hu, Y.~Jiang, C.-W. Fu, and P.-A. Heng, ``Mask-shadowgan: Learning to remove
  shadows from unpaired data,'' in \emph{Proceedings of the IEEE/CVF
  International Conference on Computer Vision}, 2019, pp. 2472--2481.

\bibitem{liu2021LG}
Z.~Liu, H.~Yin, Y.~Mi, M.~Pu, and S.~Wang, ``Shadow removal by a
  lightness-guided network with training on unpaired data,'' \emph{IEEE
  Transactions on Image Processing}, vol.~30, pp. 1853--1865, 2021.

\bibitem{liu2021G2R}
Z.~Liu, H.~Yin, X.~Wu, Z.~Wu, Y.~Mi, and S.~Wang, ``From shadow generation to
  shadow removal,'' in \emph{Proceedings of the IEEE/CVF Conference on Computer
  Vision and Pattern Recognition}, 2021, pp. 4927--4936.

\bibitem{gao2022towards}
J.~Gao, Q.~Zheng, and Y.~Guo, ``Towards real-world shadow removal with a shadow
  simulation method and a two-stage framework,'' in \emph{Proceedings of the
  IEEE/CVF Conference on Computer Vision and Pattern Recognition}, 2022, pp.
  599--608.

\bibitem{richter2016playing}
S.~R. Richter, V.~Vineet, S.~Roth, and V.~Koltun, ``Playing for data: Ground
  truth from computer games,'' in \emph{European conference on computer
  vision}.\hskip 1em plus 0.5em minus 0.4em\relax Springer, 2016, pp. 102--118.

\bibitem{sidorov2019conditional}
O.~Sidorov, ``Conditional gans for multi-illuminant color constancy: Revolution
  or yet another approach?'' in \emph{Proceedings of the IEEE/CVF Conference on
  Computer Vision and Pattern Recognition Workshops}, 2019, pp. 0--0.

\bibitem{UNet2015}
O.~Ronneberger, P.~Fischer, and T.~Brox, ``{U-Net}: Convolutional networks for
  biomedical image segmentation,'' in \emph{International Conference on Medical
  Image Computing and Computer Assisted Intervention (MICCAI)}, 2015, pp.
  234--241.

\bibitem{ilg2017flownet}
E.~Ilg, N.~Mayer, T.~Saikia, M.~Keuper, A.~Dosovitskiy, and T.~Brox, ``Flownet
  2.0: Evolution of optical flow estimation with deep networks,'' in
  \emph{Proceedings of the IEEE conference on computer vision and pattern
  recognition}, 2017, pp. 2462--2470.

\bibitem{dosovitskiy2020image}
A.~Dosovitskiy, L.~Beyer, A.~Kolesnikov, D.~Weissenborn, X.~Zhai,
  T.~Unterthiner, M.~Dehghani, M.~Minderer, G.~Heigold, S.~Gelly \emph{et~al.},
  ``An image is worth 16x16 words: Transformers for image recognition at
  scale,'' \emph{arXiv preprint arXiv:2010.11929}, 2020.

\bibitem{zhang2018image}
Y.~Zhang, K.~Li, K.~Li, L.~Wang, B.~Zhong, and Y.~Fu, ``Image super-resolution
  using very deep residual channel attention networks,'' in \emph{Proceedings
  of the European conference on computer vision (ECCV)}, 2018, pp. 286--301.

\bibitem{charbonnier1994two}
P.~Charbonnier, L.~Blanc-Feraud, G.~Aubert, and M.~Barlaud, ``Two deterministic
  half-quadratic regularization algorithms for computed imaging,'' in
  \emph{Proceedings of 1st International Conference on Image Processing},
  vol.~2.\hskip 1em plus 0.5em minus 0.4em\relax IEEE, 1994, pp. 168--172.

\bibitem{rubinstein2004cross}
R.~Y. Rubinstein and D.~P. Kroese, \emph{The cross-entropy method: a unified
  approach to combinatorial optimization, Monte-Carlo simulation, and machine
  learning}.\hskip 1em plus 0.5em minus 0.4em\relax Springer, 2004, vol. 133.

\bibitem{yang2020fda}
Y.~Yang and S.~Soatto, ``Fda: Fourier domain adaptation for semantic
  segmentation,'' in \emph{Proceedings of the IEEE/CVF Conference on Computer
  Vision and Pattern Recognition}, 2020, pp. 4085--4095.

\bibitem{frigo1998fftw}
M.~Frigo and S.~G. Johnson, ``Fftw: An adaptive software architecture for the
  fft,'' in \emph{Proceedings of the 1998 IEEE International Conference on
  Acoustics, Speech and Signal Processing, ICASSP'98 (Cat. No. 98CH36181)},
  vol.~3.\hskip 1em plus 0.5em minus 0.4em\relax IEEE, 1998, pp. 1381--1384.

\bibitem{novak1992anatomy}
C.~L. Novak, S.~A. Shafer \emph{et~al.}, ``Anatomy of a color histogram.'' in
  \emph{CVPR}, vol.~92, 1992, pp. 599--605.

\bibitem{loshchilov2016sgdr}
I.~Loshchilov and F.~Hutter, ``Sgdr: Stochastic gradient descent with warm
  restarts,'' \emph{arXiv preprint arXiv:1608.03983}, 2016.

\bibitem{chatterjee1986influential}
S.~Chatterjee and A.~S. Hadi, ``Influential observations, high leverage points,
  and outliers in linear regression,'' \emph{Statistical science}, pp.
  379--393, 1986.

\end{thebibliography}

\begin{IEEEbiography}[{\includegraphics[width=1in,height=1.25in,clip,keepaspectratio]{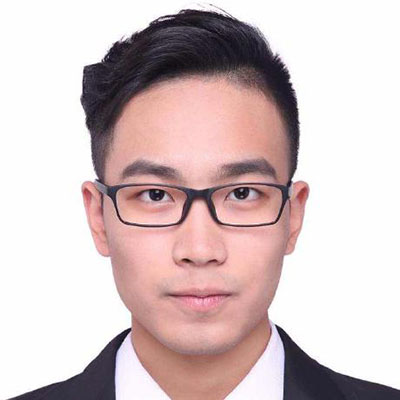}}]{Zhihao Chen}
received the B.S. degree in software engineering from the Tianjin University, Tianjin, China, in 2017. He is currently pursuing the Ph.D. degree in College of Intelligence and Computing from the Tianjin University, China. His research interests include computer vision and deep learning, specifically for shadow detection shadow removal, and medical image segmentation.
\end{IEEEbiography}

\begin{IEEEbiography}[{\includegraphics[width=1in,height=1.25in,clip,keepaspectratio]{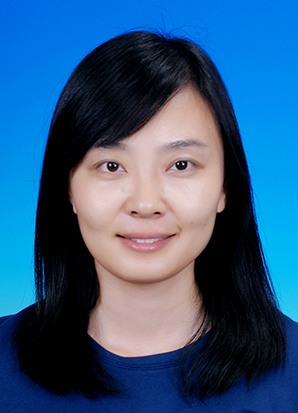}}]{Liang Wan}
is a full Professor in the College of Intelligence Computing, and deputy director of Medical College, Tianjin University, P. R. China. She obtained a Ph.D. degree in computer science and engineering from The Chinese University of Hong Kong in 2007, and worked as a PostDoc Research Associate/Fellow at City University of Hong Kong from 2007 to 2011. Her current research interests focus on image processing and computer vision, including image segmentation, low-level image restoration, and medical image analysis.
\end{IEEEbiography}

\begin{IEEEbiography}[{\includegraphics[width=1in,height=1.25in,clip,keepaspectratio]{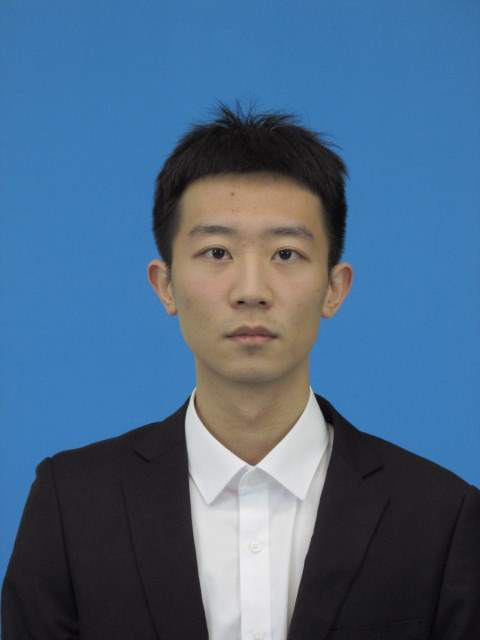}}]{Yefan Xiao}
received the bachelor degree in software engineering from the Tianjin University, Tianjin, China, in 2020. He is currently pursuing the master degree in Tianjin University. His research interests include computer vision and deep learning, specifically for image polyp segmentation and video semantic segmentation.
\end{IEEEbiography}

\begin{IEEEbiography}[{\includegraphics[width=1in,height=1.25in,clip,keepaspectratio]{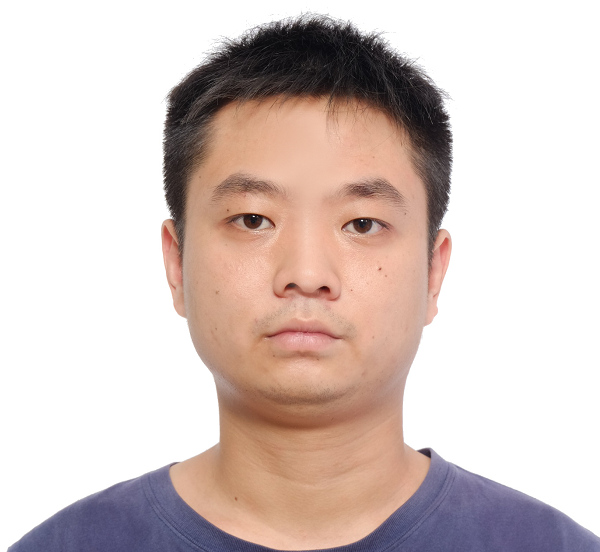}}]{Lei Zhu}
received the Ph.D. degree from the Department
of Computer Science and Engineering, The
Chinese University of Hong Kong. He is currently
working as an Assistant Professor with the ROAS
Thrust, HKUST (GZ), and also an affiliated Assistant
Professor in ECE with HKUST. Before that,
he was a Postdoctoral Researcher at DAMTP, University
of Cambridge. His research interests include
computer graphics, computer vision, medical image
processing, and deep learning.
\end{IEEEbiography}

\begin{IEEEbiography}[{\includegraphics[width=1in,height=1.25in,clip,keepaspectratio]{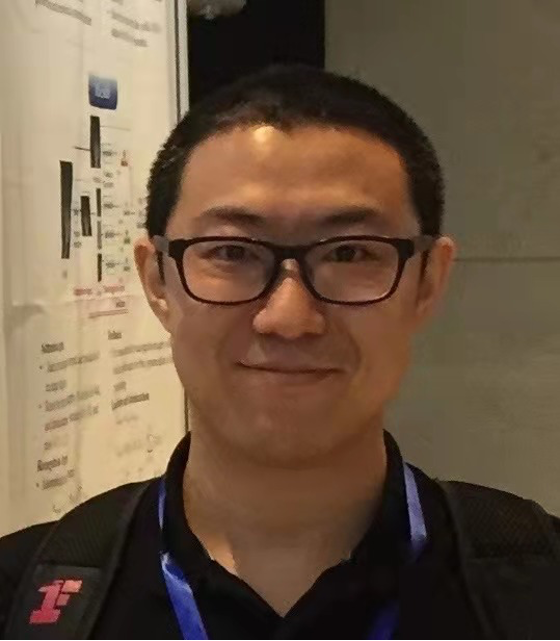}}]{Huazhu Fu}
(SM'18) is a senior scientist at Institute of High Performance Computing (IHPC), A*STAR, Singapore. He received his Ph.D. from Tianjin University in 2013. Previously, he was a Research Fellow (2013-2015) at NTU, Singapore, a Research Scientist (2015-2018) at I2R, A*STAR, Singapore, and a Senior Scientist (2018-2021) at Inception Institute of Artificial Intelligence, UAE. His research interests include computer vision, AI in healthcare, and trustworthy AI. He received the Best Paper Award from ICME 2021. He has served as the AE of IEEE TMI, IEEE TNNLS, and IEEE JBHI, AC/Senior-PC for MICCAI, IJCAI, and AAAI. He is also a Member of the IEEE BISP TC.
\end{IEEEbiography}

\newpage
\vfill

\end{document}